\documentclass[Afour,sageh,times]{sagej}

\usepackage{moreverb,url}
\usepackage[colorlinks,bookmarksopen,bookmarksnumbered,citecolor=red,urlcolor=red,draft]{hyperref}

\newcommand\BibTeX{{\rmfamily B\kern-.05em \textsc{i\kern-.025em b}\kern-.08em
T\kern-.1667em\lower.7ex\hbox{E}\kern-.125emX}}

\usepackage{amsmath}
\usepackage{mathtools}
\usepackage{algorithm}
\usepackage{array}
\usepackage[noend]{algpseudocode}
\usepackage{amssymb} 
\usepackage{graphicx}
\usepackage[font=footnotesize]{subcaption}
\usepackage[font=footnotesize]{caption}
\usepackage{amssymb}
\usepackage{url}
\usepackage{color}
\usepackage{xcolor,colortbl}
\usepackage{ifthen}
\usepackage{xspace}
\usepackage{multirow}
\usepackage[normalem]{ulem}
\usepackage{tikz}
\usetikzlibrary{arrows.meta}
\usetikzlibrary{shapes}

\usetikzlibrary{positioning}

\usepackage{soul}
\sethlcolor{initBlueHighlight}

\usepackage{makecell}

\def\volumeyear{2023}

\begin{document}

\runninghead{Holladay, Lozano-P\'{e}rez and Rodriguez}

\title{Robust Planning for Multi-stage Forceful Manipulation}

\author{Rachel Holladay\affilnum{1}, Tom\'{a}s Lozano-P\'{e}rez\affilnum{1} and Alberto Rodriguez\affilnum{2}}

\affiliation{\affilnum{1}Computer Science and Artificial Intelligence Laboratory, MIT\\
\affilnum{2}Mechanical Engineering Department, MIT}

\corrauth{Rachel Holladay \\
CSAIL, MIT, Cambridge, MA 02139, USA.}

\email{rhollada@mit.edu}

\newcommand{\eref}[1]{(\ref{#1})} 
\newcommand{\sref}[1]{Sec. \ref{#1}} 
\newcommand{\apref}[1]{Appendix \ref{#1}} 
\newcommand{\figref}[1]{Fig.\ref{#1}} 
\newcommand{\tref}[1]{Table \ref{#1}} 
\newcommand{\aref}[1]{Algorithm \ref{#1}} 
\newcommand{\lref}[1]{Line \ref{#1}} 
\renewcommand*\rmdefault{ppl}
\setlength{\textfloatsep}{5pt}

\definecolor{effectBrown}{RGB}{115, 58, 27}
\definecolor{searchYellow}{RGB}{239, 187, 26}
\definecolor{samplerGreen}{RGB}{56, 118, 29}
\definecolor{initBlueHighlight}{RGB}{149, 202, 245}
\definecolor{initBlue}{RGB}{18, 117, 197}
\definecolor{actionGrey}{gray}{0.95}

\definecolor{addedKinematic}{gray}{1}
\definecolor{addedFM}{gray}{0.9}
\definecolor{addedTask}{gray}{0.82}

\definecolor{rGray}{RGB}{171, 171, 171}
\definecolor{rBlue}{RGB}{134, 134, 183}
\definecolor{rRed}{RGB}{200, 98, 98}
\newcommand{\markerGray}{\raisebox{0.5pt}{\tikz{\node[draw,scale=0.4,regular polygon, regular polygon sides=4,fill=rGray](){};}}}
\newcommand{\markerBlue}{\raisebox{0.5pt}{\tikz{\node[draw,scale=0.4,regular polygon, regular polygon sides=4,fill=rBlue](){};}}}
\newcommand{\markerRed}{\raisebox{0.5pt}{\tikz{\node[draw,scale=0.4,regular polygon, regular polygon sides=4,fill=rRed](){};}}}

\definecolor{rLow}{RGB}{23, 204, 23}
\definecolor{rMedium}{RGB}{23, 159, 23}
\definecolor{rHigh}{RGB}{20, 98, 20}
\newcommand{\markerLow}{\raisebox{0.5pt}{\tikz{\node[draw,scale=0.4,regular polygon, regular polygon sides=4,fill=rLow](){};}}}
\newcommand{\markerMedium}{\raisebox{0.5pt}{\tikz{\node[draw,scale=0.4,regular polygon, regular polygon sides=4,fill=rMedium](){};}}}
\newcommand{\markerHigh}{\raisebox{0.5pt}{\tikz{\node[draw,scale=0.4,regular polygon, regular polygon sides=4,fill=rHigh](){};}}}

\newcommand{\etal}{et al.}

\newboolean{include-notes}
\setboolean{include-notes}{true}
\newcommand{\rhnote}[1]{\ifthenelse{\boolean{include-notes}}%
 {\textcolor{red}{\textbf{RH: #1}}}{}}
\newcommand{\rhtodo}[1]{\ifthenelse{\boolean{include-notes}}%
 {\textcolor{blue}{\textbf{RH DO: #1}}}{}}
\newcommand{\tlpnote}[1]{\ifthenelse{\boolean{include-notes}}%
 {\textcolor{orange}{\textbf{From TLP: #1}}}{}}
\newcommand{\arnote}[1]{\ifthenelse{\boolean{include-notes}}%
 {\textcolor{green}{\textbf{AR: #1}}}{}}

\newcommand{\variable}[1]{$\mathit{#1}$}
\newcommand{\constant}[1]{$\mathbf{#1}$}

\begin{abstract}
Multi-step forceful manipulation tasks, such as opening a push-and-twist childproof bottle, require a robot to make various planning choices that are substantially impacted by the requirement to exert force during the task. 
The robot must reason over discrete and continuous choices relating to the sequence of actions, such as whether to pick up an object, and the parameters of each of those actions, such how to grasp the object. 
To enable planning and executing forceful manipulation, we augment an existing task and motion planner with constraints that explicitly consider torque and frictional limits, captured through the proposed forceful kinematic chain constraint. 
In three domains, opening a childproof bottle, twisting a nut and cutting a vegetable, we demonstrate how the system selects from among a combinatorial set of strategies.
We also show how cost-sensitive planning can be used to find strategies and parameters that are robust to uncertainty in the physical parameters.
\end{abstract}


\maketitle


\section{Introduction}
\label{sec:intro}


Our goal is to enable robots to plan and execute {\em forceful manipulation tasks} such as cutting a vegetable, opening a push-and-twist childproof bottle, and twisting a nut.  
While all tasks that involve contact are technically forceful, we refer to forceful manipulation tasks as those where the ability to generate and transmit the necessary forces to objects and their environment is an active limiting factor which must be reasoned over and planned for. 
Respecting these limits might require a planner, for example, to prefer a robot configuration that can exert more force or a grasp on an object that is more stable. 

Forceful operations, as defined by \cite{chen2019manipulation}, are the exertion of a wrench (generalized force/torque) at a point on an object.  
These operations are intended to be quasi-statically stable, i.e. the forces are always in balance and produce relatively slow motions, and will generally require some form of fixturing to balance the applied wrenches.  
For example, to open a push-and-twist childproof bottle 
the robot must exert a downward force on the cap while applying a torque along the axis of the bottle.
The robot must be in a configuration that allows it to apply enough force to accomplish the task, and also securing the bottle to prevent it from moving during the task. 

\begin{figure*}[t]
\centering
    \includegraphics[width=0.9\textwidth]{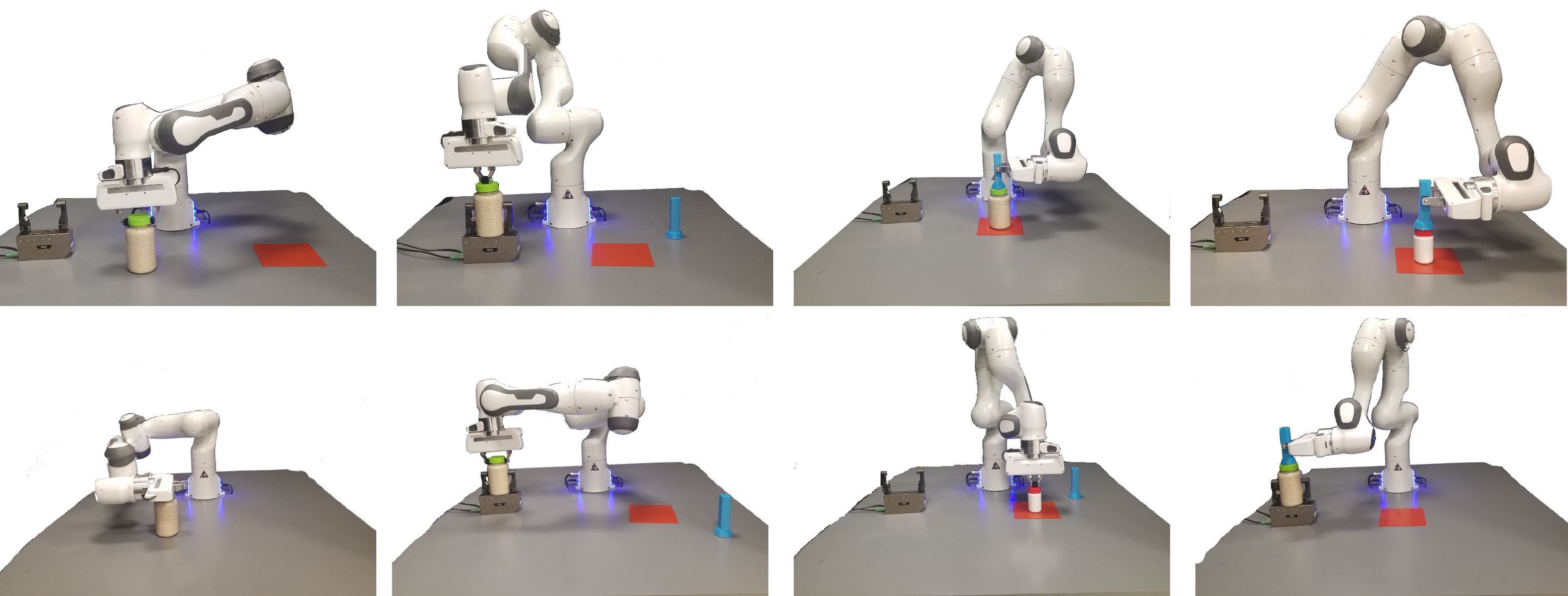}
\caption{Opening a childproof bottle involves executing a downward-push and twist on the cap, while fixturing the bottle. Our system can reason over a combinatorial number of strategies to accomplish this forceful manipulation task, including push-twisting with various parts of its end effector, push-twisting with a tool (in blue), fixturing with a vise (in grey), fixturing against the table, or fixturing against a high-friction rubber mat (in red).}
\label{fig:fig1}
\end{figure*}

To accomplish these complex, multi-step forceful manipulation tasks, robots need to make discrete decisions, such as, for example, whether to push on the bottle cap with the fingers, the palm or a tool, and whether to secure the bottle via frictional contact with a surface, with another gripper or with a vise. 
The robot must also make continuous decisions such as the choice of grasp pose, robot configurations and robot trajectories. 
Critically, all these decisions interact in relatively complex ways to achieve a valid task execution.

\figref{fig:fig1} illustrates that there are \textit{different strategies} for completing the task of opening a push-and-twist childproof bottle. 
Each strategy's viability depends on the robot's choices and on the environment. 
For example, a strategy that uses the friction from the table to secure the bottle, as shown in the top leftmost corner, would fail if the table can only provide a small amount of friction.  
Instead, the robot would need to find a significantly different strategy, such as securing, or fixturing, the bottle via a vise, as shown in the top center-left.

Choosing a strategy corresponds to making some of the aforementioned discrete decisions, \textit{e.g.} deciding how to fixture the bottle. 
We define strategies as sequences of parameterized high-level actions.  
Each action is implemented as a controller parameterized by a set of constrained discrete and continuous values, such as robot configurations, objects, grasp poses, trajectories, etc.  
Our goal is to find both a sequence of high-level actions (a strategy) and parameter values for those actions, all of which satisfy the various constraints on the robot's motions. 

To produce valid solutions for a wide range of object and environment configurations, the robot must be able to consider a wide range of strategies. 
As discussed above, small changes, such as decreasing a friction coefficient, may necessitate an entirely new strategy. 
In a different environment, the robot may need to first move some blocking object out of the way (\figref{fig:removeBlock}) or relocate an object in order to achieve a better grasp. 
Approaches that attempt to explicitly encode strategies in the form of a policy, \textit{e.g.} via a finite state machine or a fixed action sequence, will generally fail to capture the full range of feasible strategies~\citep{michelman1994forming,holladay2019force}. 
Methods that attempt to learn such a policy will need a very large number of interactions to explore this rich and highly-constrained solution space. 

We propose addressing forceful manipulation problems by planning over the combinatorial set of discrete/continuous choices.
We use an existing \textit{task and motion planning} (TAMP) system, PDDLStream, which reduces this type of hybrid discrete/continuous planning problem to a sequence of discrete planning problems via focused sampling of the continuous and discrete parameters~\citep{garrett2020pddlstream}.

To enable this approach to forceful manipulation, we introduce force-related constraints: the requirement to fixture objects and the \textit{forceful kinematic chain}. 
The forceful kinematic chain constraint captures whether the robotic and frictional joints of the kinematic chain are ``strong'' enough to exert the forces and torques to perform forceful operations. 
These constraints are integrated into the TAMP framework.

Furthermore, we enable the planner to choose strategies that are \textit{robust to uncertainty} by formulating this as cost-sensitive planning, where the cost of an action is tied to its probability of success in open-loop execution, given perturbations in parameters for the force-based constraints.


Our paper makes the following contributions: 
\begin{itemize}
    \item Characterize forceful manipulation in terms of constraints on forceful kinematic chains
    \item Generate multi-step plans that obey force- and motion-related constraints using an existing TAMP framework.
    \item Formulate finding plans that are robust to uncertainty in the physical parameters of forceful kinematic chains as cost-sensitive planning. 
    \item Demonstrate planning and robust planning for forceful manipulation in three domains (opening a childproof bottle, twisting a nut on a bolt, cutting a vegetable) 
\end{itemize}

This paper is an extension of \citet{holladay2021planning} and the additional contributions beyond that work are: a new domain (cutting a vegetable), additional demonstrations relating to robust planning and more detailed descriptions of the planning framework and domain specifications.

\section{Related Work}
\label{sec:related_work} 

This paper focuses on enabling force-based reasoning in planning multi-step manipulation tasks.
In this section we review various strategies for incorporating force-based reasoning and force-related constraints across various levels of planning: from single step actions to fixed action sequences to task and motion planning. 

\subsection{Force-Reasoning in Single Actions}

Several papers have considered force requirements for generating specialized motions. 
Gao \etal~use learning from demonstration to capture ``force-relevant skills'', defined as a desired position and velocity in task space, along with an interaction wrench and task constraint~\citep{gao2019learning}.
Berenson \etal~incorporated a torque-limiting constraint into a sample-based motion planning to enable manipulation of a heavy object~\citep{berenson2009manipulation}. 
These papers consider force constraints when generating individual actions, while we consider force constraints over a sequence of actions. 

One important category of reasoning with respect to force actions is stabilizing, or fixturing an object. 
The goal of fixturing is to fully constrain an object or part, while enabling it to be accessible~\citep{asada1985kinematics}.
Fixture planning often relies on a combination of geometric, force and friction analyses~\citep{hong1991fixture}.
There are various methods of fixturing including using clamps~\citep{mitsioni2019data}, custom jigs~\citep{levi2022simjig} or using another robot to directly grasp or grasp via tongs~\citep{watanabe2013cooking,stuckler2016mobile,zhang2019leveraging}. 
Additionally some strategies, such as fixtureless fixturing and shared grasping, rely on friction and environmental contacts to fixture without additional fixtures or tools~\citep{chavan2018regrasping,hou2020manipulation}. 
In this paper we consider fixturing via grasping and environmental contacts.

\subsection{Force-Reasoning in Sequences of Actions}

Several papers have considered reasoning over forces across multi-step interactions. 
Chen \etal~define ``forceful operations'' as a 6D wrench $f$ applied at a pose $p$ with respect to a target object~\citep{chen2019manipulation}. 
In this paper, we adopt their definition of forceful operations to characterize the type of interactions our system plans for. 
Chen \etal~focus on searching over environmental and robot contacts to stabilize an object while a human applies a series of forceful operations. 
In our work, a robot must both stabilize the object and perform the forceful operation. 

Manschitz \etal~termed ``sequential forceful interaction tasks'' as those characterized by point-to-point motions and an interaction where the robot must actively apply a wrench~\citep{manschitz2020learning}.
The method first learns, from demonstrations, a set of movement primitives, which are then sequenced by learning a mapping from feature vectors to activation of the primitives.
The goal of their work is enable the robot to reproduce the demonstrations and therefore the sequencing is pre-scripted. 

Michelman and Allen formalize opening a childproof bottle via a finite state machine, where the robot iteratively rotates the cap, while pressing down, and then attempts to lift it until the cap moves~\citep{michelman1994forming}.
If the cap is not yet free, the robot continues to rotate the cap. 
In their formulation, the bottle is fixtured and some of the continuous parameters, such as the grasp, are fixed. 

Holladay \etal~frames tool use as a constraint satisfaction problem where the planner must choose grasps, arm paths and tool paths subject to force and kinematic constraints such as joint torque limits, grasp stability, environment collisions, etc.~\citep{holladay2019force}.  
The planner outputs a fixed sequence of actions and thus the force and kinematic constraints do not impact the sequence of actions, i.e. the choice of strategy. 
In contrast, in this paper, we are interested in searching over various possible strategies. 

\subsection{Multi-step Planning with Constraints}

Solving for a sequence of actions parameterized by constrained and continuous values lies at the heart of multi-modal motion planning (MMMP)~\citep{hauser2010multi,hauser2011randomized} and task and motion planning (TAMP)~\citep{garrett2020integrated}. 
MMMP plans motions that follow kinematic modes, e.g. moving through free space or pushing an object, and motions that switch between discrete modes, e.g. grasping or breaking contact, where each mode is a submanifold of configuration space. 

TAMP extends MMMP by incorporating non-geometric state variables and a structured action representation that supports efficient search~\citep{gravot2005asymov,plaku2010sampling,kaelbling2011hierarchical,srivastava2014combined,toussaint2015logic,dantam2016incremental,garrett2017sample}. 

Most TAMP algorithms, although not all~\citep{toussaint2020describing}, have focused on handling collision and kinematic constraints. 
This paper focuses on integrating force-based constraints with an existing TAMP framework, PDDLStream~\citep{garrett2020pddlstream}, which we discuss in \sref{sec:pddlstream}.

Most similar to our work, Toussaint \etal~formulate force-related constraints that integrate into a trajectory-optimization framework (LGP) for manipulation tasks~\citep{toussaint2020describing}.  
While LGP can search over strategies, in the aforementioned paper the strategy was provided and fixed. 
While Toussaint \etal~take a more generic approach to representing interaction which can capture dynamic manipulation, their use of 3D point-of-attack (POA) to represent transmitted 6D wrenches,
fails short of representing the frictional constraints of the contact patches.

Levihn and Stilmann present a planner that reasons over which combination of objects in the environment will yield the appropriate mechanical advantage for unjamming a door~\citep{levihn2014using}. 
The type of the door directly specifies which strategy to use (lever or battering ram) and the planner considers the interdependencies of force-based and geometric-based decisions for each application.
The planer is specific to the domain and does not provide a general framework for planning with force constraints. 

Several other systems consider forces either as a feasibility constraint or as a cost. 
In assessing feasibility for assembly plans, \cite{lee1993physical} account for the amount of force required to connect two pieces in an assembly. 
\citet{akbari2015task} focus on incorporating ``physics-based reasoning'' in a TAMP system that sequences push and move actions by formulating action costs with respect to power consumed and forces applied. 
Again, each of these planners present a domain-specific approach to considering force.

In this work, we consider generating plans that are robust to state uncertainty, with a particular focus on physical properties of the objects.
Several TAMP frameworks approach uncertainty by planning in belief space, the space of probability distributions over underlying world states~\citep{kaelbling2013integrated,hadfield2015modular,LIS267}. 
These planners generate action sequences that can also involve information-gathering sensing actions. 
In contrast, our system executes open-loop plans and focuses on uncertainty that impacts the force-related constraints. 

\section{Problem Domain}
\label{sec:domains}

We define \textit{forceful manipulation} as a class of multi-step manipulation tasks that involve reasoning over and executing forceful operations. 
Drawing from \citet{chen2019manipulation}, a forceful operation is a robot action where the robot exerts a 6D force/torque wrench ($[f_{x}, f_{y}, f_{z}, t_{x}, t_{y}, t_{z}]$) on an object or the environment.
In addition to the geometric constraints (such as collision-free trajectories) that characterize many multi-step manipulation tasks, forceful manipulation tasks are also characterized by constraints relating to the ability to exert wrenches. 

Specifically, we constrain that the robot, including any grasped object, must be ``strong'' or ``stable'' enough to exert the forceful operation, i.e. the robotic system must be able to exert the desired wrench of the forceful operation without experiencing excessive force errors or undesired slip. 
We additionally constrain that any object that the forceful operation is acting on must be secured, or fixtured, in a way that prevents its motion.  

Our aim is to perform forceful manipulation tasks in a wide range of environments, where there is variation in the number, type, poses and physical parameters, such as masses and friction coefficients, of the objects. 
We also characterize \textit{robust} planning for forceful manipulation tasks as generating plans whose success is robust to uncertainty in these physical parameters.

To ground our work in concrete problems, we consider three example tasks within the forceful manipulation domains: (1) opening a childproof bottle (2) twisting a nut on a bolt and (3) cutting a vegetable. 
For each domain, we define the forceful operation(s) that represents the task (also called the task wrench(es)), strategies for exerting those forceful operation(s) and what object must be fixtured, 
We then discuss various methods of fixturing objects. 

\begin{figure}[t!]
  \begin{subfigure}[t]{0.58\columnwidth}
    \begin{tikzpicture}[box/.style={rectangle, align=center}]
       \node     {\includegraphics[width=0.35\textwidth]{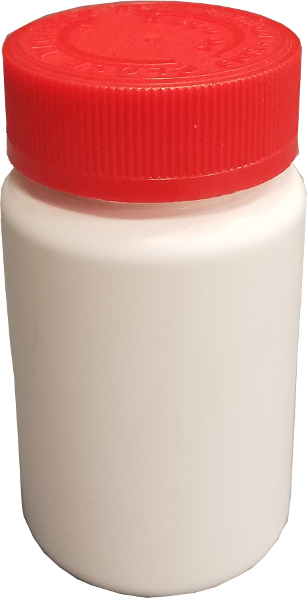}};
       \draw[-latex, gray, line width=0.04cm] (0, 1.45) --+(0, 1) node[above] {{\LARGE $z$}};
       \draw[-latex, gray, line width=0.04cm] (0, 1.45) --+(1.25, 0) node[right] {{\LARGE $x$}};
       \draw[-latex, gray, line width=0.04cm] (0, 1.45) --+(-0.7, -0.7) node[left] {{\LARGE $y$}};
       \draw[-{Latex[length=10pt]}, line width=0.1cm] (0, 1.45) --+(0, -1.15);
       \draw[-{Latex[length=10pt]}, line width=0.1cm, rotate=0] (-0.35, 1.7) arc [start angle=-240, end angle=70, x radius=0.8cm, y radius=0.3cm];
       \draw[line width=0.05cm] (0, -1.8) --+(0, 0.15)        (-0.5, -1.8) --+(1, 0);
       \draw[line width=0.05cm] (-0.5, -1.8) --+(-0.15, -0.15)     (-0.16, -1.8) --+(-0.15, -0.15); 
       \draw[line width=0.05cm] ( 0.23, -1.8) --+(-0.15, -0.15)    ( 0.5, -1.8) --+(-0.15, -0.15);
    \end{tikzpicture}
    \caption{Opening Childproof Bottle} \label{fig:bottleWrench}
  \end{subfigure}
  \begin{subfigure}[t]{0.28\columnwidth}
    \begin{tikzpicture}[box/.style={rectangle, align=center}]
       \node     {\includegraphics[width=0.58\textwidth]{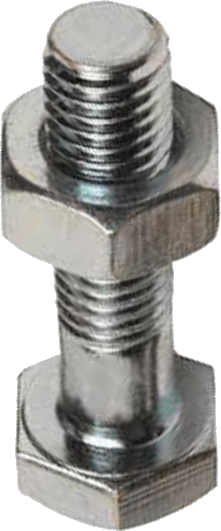}};
       \draw[-latex, gray, line width=0.04cm] (-0.05, 0.7) --+(0, 1.25) node[above] {{\LARGE $z$}};
       \draw[-latex, gray, line width=0.04cm] (-0.05, 0.7) --+(1.5, 0) node[right] {{\LARGE $x$}};
       \draw[-latex, gray, line width=0.04cm] (-0.05, 0.7) --+(-0.9, -0.9) node[left] {{\LARGE $y$}};
       \draw[-{Latex[length=10pt]}, line width=0.1cm, rotate=0] (-0.5, 1.2) arc [start angle=-240, end angle=70, x radius=1.1cm, y radius=0.6cm];
       \draw[line width=0.05cm] (0, -1.8) --+(0, 0.15)        (-0.5, -1.8) --+(1, 0);
       \draw[line width=0.05cm] (-0.5, -1.8) --+(-0.15, -0.15)     (-0.16, -1.8) --+(-0.15, -0.15);   
       \draw[line width=0.05cm] ( 0.23, -1.8) --+(-0.15, -0.15)    ( 0.5, -1.8) --+(-0.15, -0.15);
    \end{tikzpicture}
    \caption{Twisting a Nut} \label{fig:nut_task_wrench}
  \end{subfigure} \\
  \begin{subfigure}[t]{0.99\columnwidth}
    \begin{tikzpicture}[box/.style={rectangle, align=center}]
      \node   {\includegraphics[width=0.99\textwidth]{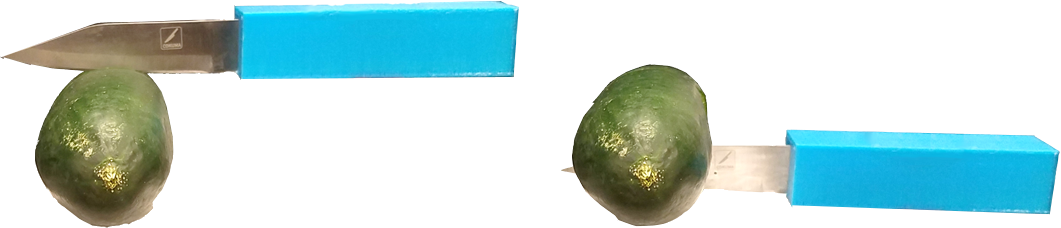}};
       \draw[-latex, gray, line width=0.04cm] (-3.35, -0.25) --+(0, 1) node[above] {{\Large $z$}};
       \draw[-latex, gray, line width=0.04cm] (-3.35, -0.25) --+(1.25, 0) node[right] {{\Large $x$}};
       \draw[-latex, gray, line width=0.04cm] (-3.35, -0.25) --+(-0.6, -0.6) node[left] {{\Large $y$}};
       \draw[-{Latex[length=8pt]}, line width=0.08cm] (-3.35, 0.3) --+(0, -1.15);
       \draw[gray, line width=0.05cm] (-3.8, -0.9) --+(1, 0);  
       \draw[line width=0.05cm] (-3.3, -1.1) --+(0, 0.15)        (-3.8, -1.1) --+(1, 0);
       \draw[line width=0.05cm] (-3.8, -1.1) --+(-0.15, -0.15)     (-3.46, -1.1) --+(-0.15, -0.15);   
       \draw[line width=0.05cm] (-3.13, -1.1) --+(-0.15, -0.15)    ( -2.8, -1.1) --+(-0.15, -0.15);

       \draw[-latex, gray, line width=0.04cm] (0.9, -0.5) --+(0, 1) node[above] {{\Large $z$}};
       \draw[-latex, gray, line width=0.04cm] (0.9, -0.5) --+(1.25, 0) node[above] {{\Large $x$}};
       \draw[-latex, gray, line width=0.04cm] (0.9, -0.5) --+(-0.6, -0.6) node[left] {{\Large $y$}};
       \draw[gray, line width=0.05cm] (0.4, -0.9) --+(1, 0);
       \draw[line width=0.05cm] (0.9, -1.1) --+(0, 0.15)        (0.4, -1.1) --+(1, 0);
       \draw[line width=0.05cm] (0.4, -1.1) --+(-0.15, -0.15)    (0.73, -1.1) --+(-0.15, -0.15);    
       \draw[line width=0.05cm] (1.06, -1.1) --+(-0.15, -0.15)   (1.4, -1.1) --+(-0.15, -0.15);
       \draw[-{Latex[length=8pt]}, line width=0.08cm] (0.9, -0.5) --+(0, -0.6);
       \draw[-{Latex[length=8pt]}, line width=0.09cm] (0.9, -0.5)--+(2, 0); 
    \end{tikzpicture}
    \caption{Cutting a vegetable, in two stages} \label{fig:cut_task_wrench}
  \end{subfigure}
\caption{Top Left: Opening a childproof bottle involves executing a push-twist on the cap, while fixturing the bottle. Top Right: Twist a nut requires exerting a torque about the nut, while fixturing the bolt. Bottom: To cut, the robot first press down vertically and then slices horizontally. The object being cut must be fixtured.} \label{fig:forcefulOperations}
\end{figure}
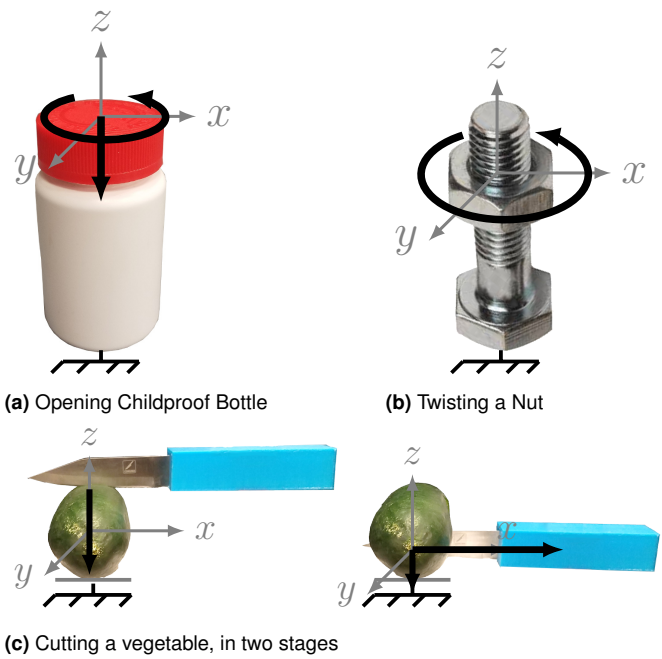

\subsection{Childproof Bottle Opening}
\label{sec:bottleDomain}

In the first domain, the objective is to open a push-and-twist childproof bottle, as introduced in \sref{sec:intro}.
We specify the push-twist, required before removing the cap, as the forceful operation of applying wrench $(0, 0, -f_{z}, 0, 0, t_{z})$ in the frame of the cap (\figref{fig:bottleWrench}), where we assume $f_{z}$ and $t_{z}$ are given.
As illustrated in \figref{fig:fig1}, the robot can apply this wrench through a variety of possible contacts: a grasp, fingertips, a palm or a grasped pusher tool. 
If using the latter three contacts, the robot can reason over applying additional downward force.  
While performing the forceful operation, the robot must fixture the bottle. 

\subsection{Nut Twisting}
\label{sec:nutTwistingDomain}

\begin{figure}[t!]
\centering
  \begin{subfigure}[t]{0.49\columnwidth}
    \includegraphics[width=\textwidth]{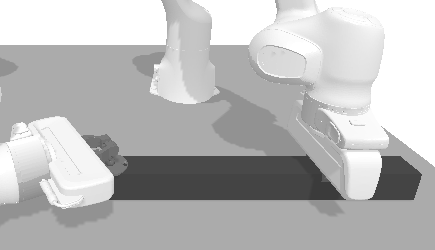}
  \end{subfigure}
  \begin{subfigure}[t]{0.49\columnwidth}
    \includegraphics[width=\textwidth]{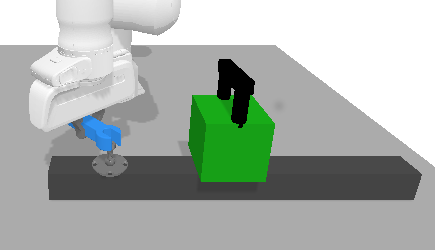}
  \end{subfigure}
\caption{To twist a nut on the bolt, the robot can use either its fingers or a spanner (in blue). While twisting, the robot must fixture the beam that the bolt is attached to. Here we show two fixturing strategies: using another robot to grasp the beam and weighing down the beam with a large mass (in green).} 
\label{fig:nutgrid}
\end{figure}

In the second domain, the robot twists a nut on a bolt by applying the wrench $(0, 0, 0, 0, 0, t_{z})$ in the frame of the nut (\figref{fig:nut_task_wrench}), where we assume $t_{z}$ is given.
As shown in \figref{fig:nutgrid}, the robot can make contact either through a grasp or through a grasped tool, i.e. a spanner\footnote{To avoid confusion between a wrench (the tool) and a wrench (a 6D force-torque) we use the British term ``spanner'' to refer to the tool.}. 
When twisting the nut, the robot must also fixture the beam holding the bolt. 

We do not consider the more general task of twisting a nut \textit{until} it is tight, which would require a feedback loop. 

\subsection{Vegetable Cutting}
\label{sec:cuttingDomain}

\begin{figure}[t!]
\centering
    \includegraphics[width=0.48\columnwidth]{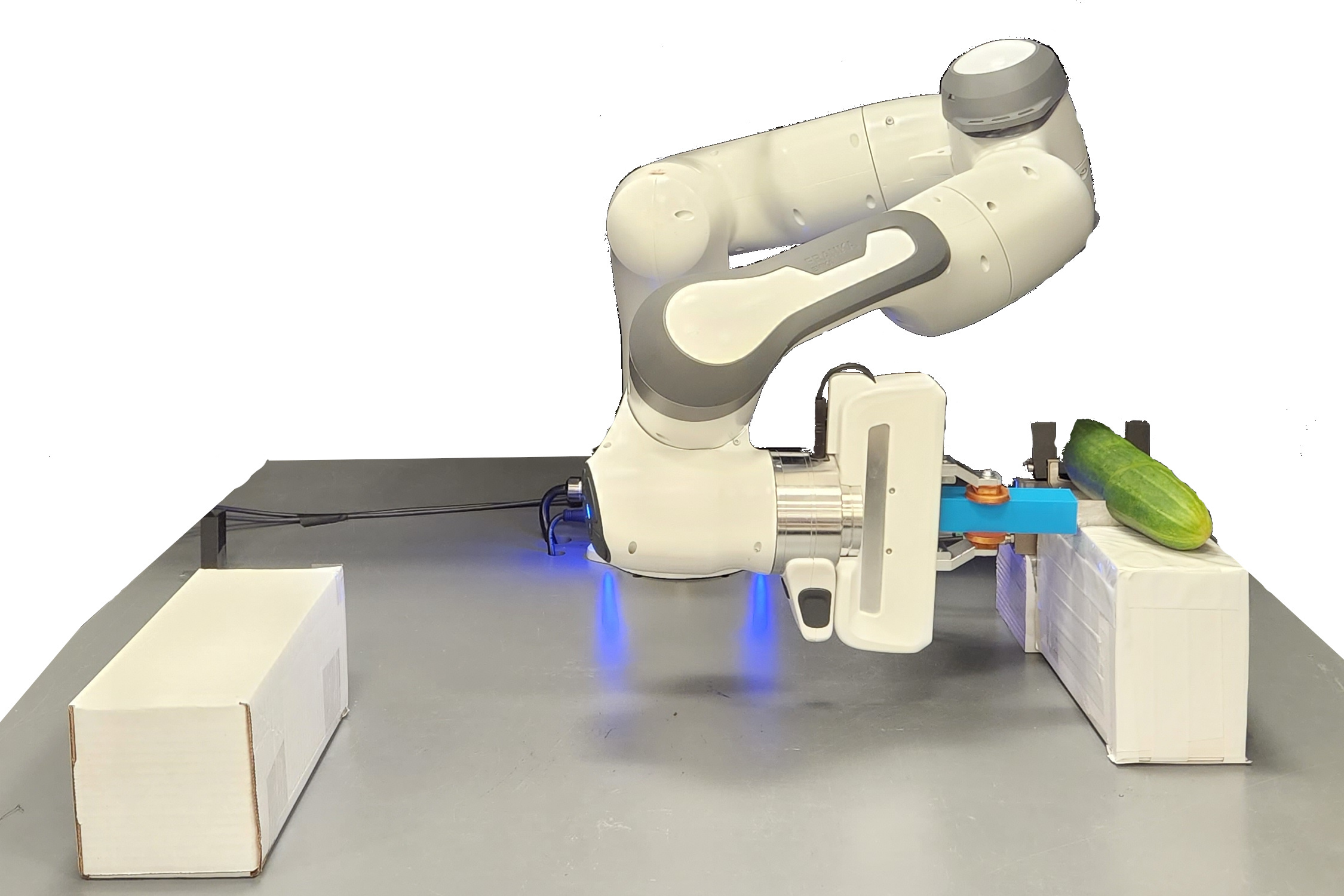}
    \includegraphics[width=0.48\columnwidth]{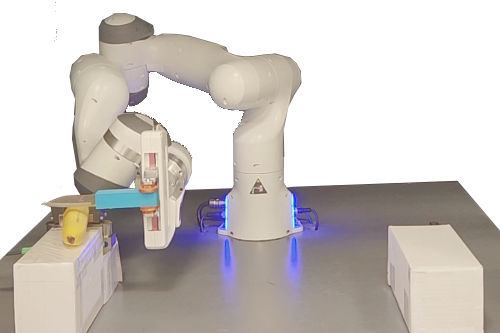}
\caption{The robot uses a knife to cut an object, while fixturing the object. Here a vise is used to fixture a cucumber (left) and a banana (right) while they are being cut with a knife (blue).}
\label{fig:cutgrid}
\end{figure}

In the third domain, the robot uses a knife to cut a vegetable (as shown in \figref{fig:cutgrid}). 
Cutting is a complex task that involves fracture, friction and changing contacts~\citep{jamdagni2019robotic,jamdagni2021robotic, mu2022physical}. 
There have been a variety of approaches to tackling this cutting-edge topic such as learning the task-specific dynamics~\citep{mitsioni2019data,zhang2019leveraging,rezaei2020learning}, developing specialized simulators~\citep{heiden2021disect}, and proposing adaptive controllers~\citep{zeng1997adaptive, long2013modeling,long2014force}.
In this work, we adopt a simplified approach to cutting. 

We take inspiration from \cite{mu2019robotic} and assume that, while being cut, the object will have negligible deformation and that dynamic effects are insignificant. 
Similarly to their proposed cutting process, we formulate cutting as a two-stage process where the knife first exerts downward force, followed by a translational slice.
Thus, this task has two forceful operations: the downward force $(0, 0, -f_{z, 0}, 0, 0, 0)$ and the translational slice $(-f_{x}, 0, f_{z, 1}, 0, 0, 0)$, as visualized in \figref{fig:cut_task_wrench}.
We assume that $f_{z, 0}$, $f_{z, 1}$ and $f_{x}$ are given. 
While cutting the object, the robot must also fixture it. 


\begin{figure*}[t!]
\centering
 \begin{tikzpicture}[ann/.style={draw=black, line width=0.03cm},]

   \node (blocker0)                                        {\includegraphics[width=0.2\textwidth]{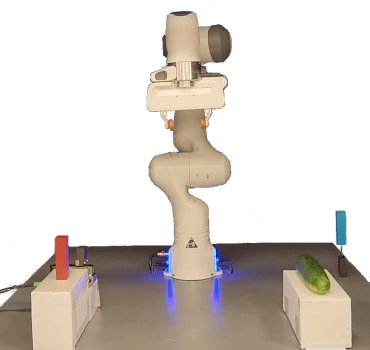}};
   \node (blocker1)    [right=of blocker0, xshift=-1.2cm]  {\includegraphics[width=0.2\textwidth]{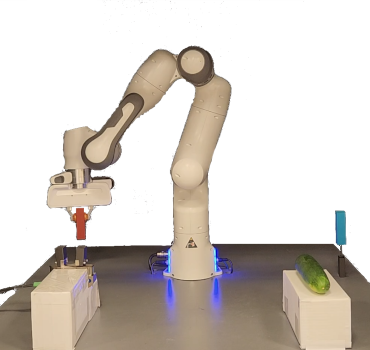}};
   \node (blocker2)    [right=of blocker1, xshift=-1.2cm]  {\includegraphics[width=0.2\textwidth]{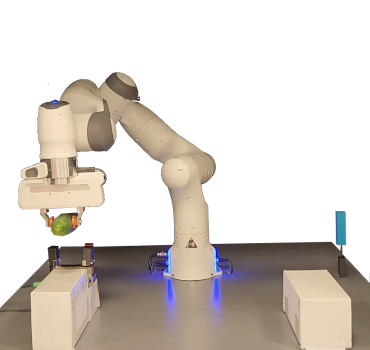}};
   \node (blocker3)    [right=of blocker2, xshift=-1.2cm]   {\includegraphics[width=0.2\textwidth]{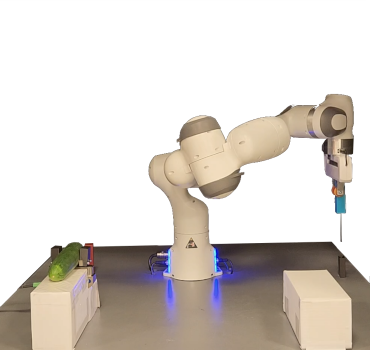}};
   \node (blocker4)    [right=of blocker3, xshift=-1.2cm]  {\includegraphics[width=0.2\textwidth]{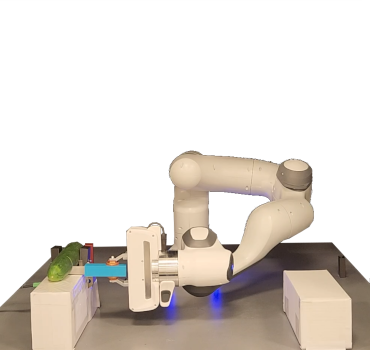}};
   
   \draw[ann] (-1.21, -0.97) rectangle ++(0.12, 0.4) node[above, xshift=-0.5cm] {Block};
   \draw[ann] (1.42, -0.66) rectangle ++(0.1, 0.34) node[above, xshift=0.2cm] {Knife};
   \draw[ann, rounded corners, rotate around={2:(-0.34, 0.54)}] (-0.34, 0.58) rectangle ++(0.73, 0.33) node[left, xshift=-0.7cm, yshift=-0.15cm] {Gripper};

   \draw[-Latex] (-0.8, -0.1) --+ (-0.1, -0.53) node[above, xshift=0.1cm, yshift=0.5cm] {Vise};
   \draw[ann] (-1.08, -0.85) rectangle ++(0.2, 0.2) node[above] {};
   \draw[-Latex] (0.78, 0.1)  --+ (0.3, -0.8) node[above, xshift=0.3cm, yshift=0.75cm] {Vegetable};
   \draw[ann, rounded corners, rotate around={30:(1.19,-1.16)}] (1.19, -1.16) rectangle ++(0.25, 0.43) node[above] {};

 \end{tikzpicture}
\caption{The goal of the robot is to cut the vegetable, using the knife (in blue). The vegetable must be fixtured, which can be achieved using the vise. However, the robot cannot secure the vegetable in the vise because a red block is preventing a collision-free placement. Our system constructs a plan where the robot first picks up the red block and places it on the table, out of the way. The robot can then pick up the vegetable and fixture it in the vise. Next, the robot grasps the knife and uses it to cut the fixtured vegetable. }
\label{fig:removeBlock}
\end{figure*}
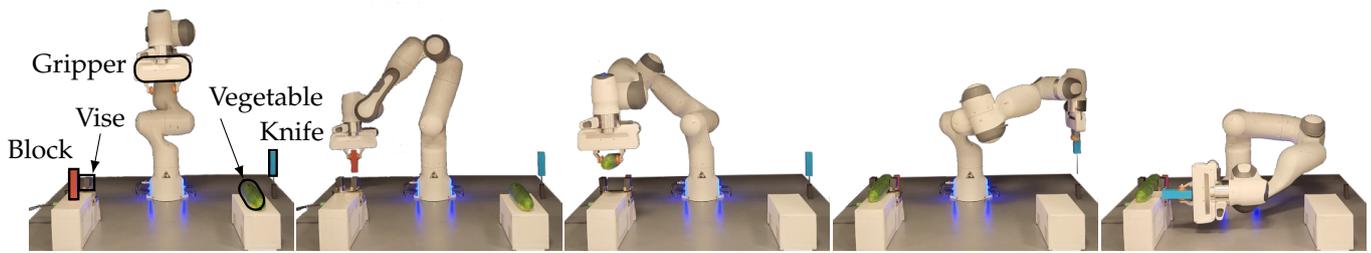

\subsection{Fixturing}
\label{sec:fixturing}

While performing any forceful operation, the robot must fixture the object it is exerting force on to prevent its motion.
There are a wide variety of ways to fixture, which are not unique to any particular domain. 
In this paper, as shown across \figref{fig:fig1}, \figref{fig:nutgrid}, \figref{fig:cutgrid}, we present several different fixturing methods such as: 
\begin{itemize} 
  \item Grasping the object with another robot
  \item Grasping the object in a vise 
  \item Weighing the object down with a heavy weight
  \item Exerting additional downward force to secure the object with friction from a surface
\end{itemize}
For the second method, in practice we use a table-mounted robot hand as the vise. 
For the last method, the frictional surface can either be the table, or higher-friction rubber mats and the robot can exert this additional downward force through various contacts: fingertips, a palm or a grasped pusher tool. 

\section{Approach}
\label{sec:approach}

Forceful manipulation tasks are characterized by constraints related to exerting force. 
We view a robotic system, composed of the robot joints, grasps and other possible frictional contacts, as a \textit{forceful kinematic chain}.
For example, in the bottom rightmost example of \figref{fig:fig1}, the robot has constructed a \textit{chain} of joints composed of the robot's actual joints, the robot's grasp on the tool and the tool's contact with the cap.

When the robot is performing a forceful operation, we can capture whether the system is strong enough to exert the task wrench by assessing if the forceful kinematic chain is maintained, i.e. if each joint is stable under the imparted wrench.
We informally use the word \textit{stable} to refer to an equilibrium of the forces and torques at all the joints.

For each class of joint, we describe a mathematical model that characterizes the set of wrenches that the joint can resist and thus the joint is stable if the imparted wrench lies within this set.
To test the constraint, the planned task wrench combined with the wrench due to gravity, is propagated through the joints of the forceful kinematic chain and each of the joints are evaluated for their stability. 

Given our domain description, there are two forceful kinematic chains when performing a forceful operation: the chain of the system exerting the task wrench and the chain of the system fixturing the object.
Returning to the bottom rightmost example of \figref{fig:fig1}, the exertion chain was described above and the fixturing chain is one joint: the vise's grasp on the bottle. 

We need a planning framework that generates multi-step manipulation plans that are flexible to a wide-range of environments and that respect various force- and motion-related constraints.
We opt to cast this as a \textit{task and motion planning (TAMP)} problem, where the robot must find a feasible strategy, or sequence of actions, to complete the task.

Each action is implemented by a parameterized controller and is associated with constraints relating the discrete and continuous parameter values that must be satisfied for the controller to achieve its desired effect.
The discrete parameters are values such as objects, regions, robot arms and the continuous parameters are values such as robot configurations, poses, paths, wrenches, etc. 

Solving a TAMP problem corresponds to finding a valid sequence of actions and finding the discrete and continuous parameters of those actions that satisfy the constraints. 
These two problems are tightly connected, since the force- and motion-related constraints impact whether it's possible to find a valid sequence of actions. 

As an illustration of this connection, we return back to the example mentioned in \sref{sec:intro}, where the table top does not provide enough friction to fixture the bottle. 
This corresponds to a forceful kinematic chain where it is impossible to find a set of parameters to make it stable.
In this case, the planner searches for a new set of actions, such as picking and placing the bottle into a vise, in order to create a different, stable forceful kinematic chain. 

As another example, \figref{fig:removeBlock} shows the robot completing a sequence of actions in order to cut a vegetable while it is fixtured in a vise. 
Since the red block on the vise prevents the robot from directly placing the vegetable in the vise, the robot constructs a plan to move the red block out of the way before fixturing and cutting the vegetable. 

Both of these examples illustrate that the interleaved constraint evaluation and action search in the TAMP framework is critical to enabling the planner to solve the tasks in a variety of environments.  

In order for plans to reliably succeed in a variety of environments, the robot must also be able to account for uncertainty in the world.
In this work, we focus on uncertainty in the physical parameters of the stability models used to evaluate the forceful kinematic chain constraint. 
For example, if the stability of a grasp used during a forceful operation is dependent on precise value of a friction coefficient, this choice of grasp is not very robust to uncertainty. 

In order to find plans that are \textit{robust}, we use cost-sensitive planning, where the planner searches for a plan whose cost is below a user-defined cost threshold.
We relate the cost of an action to the probability that the forceful kinematic chain involved in the action is stable under uncertainty in the physical parameters. 
This formulation allows the planner to reason over which strategies and which parameters of those strategies lead to more robust, reliable execution. 

In this work we assume a quasi-static physics model and, as input, are given geometric models of the robot, the objects and the environment along with the poses of each object.
Estimates of physical parameters, such as the object's mass and center of mass, and friction coefficients, are known.
In robust forceful manipulation, we relax the need for exact estimates of physical parameters, instead using ranges.

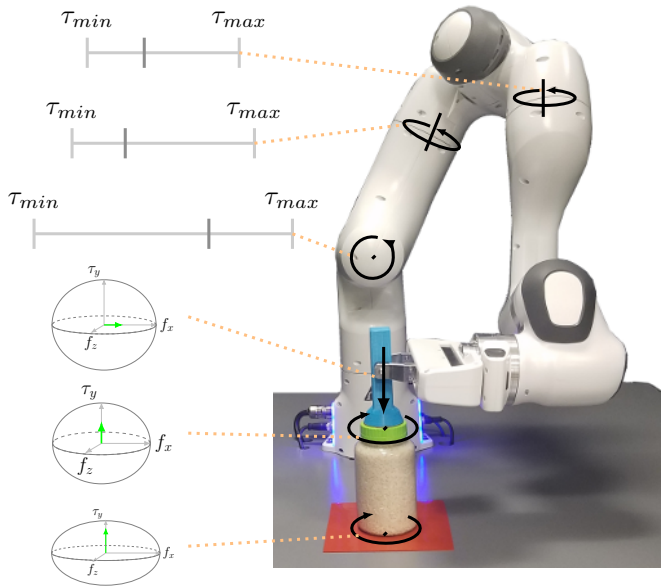
\begin{figure}[t!]
 \begin{tikzpicture}[box/.style={rectangle, align=center}]

   \node (robotPic)                      {\includegraphics[width=0.6\columnwidth]{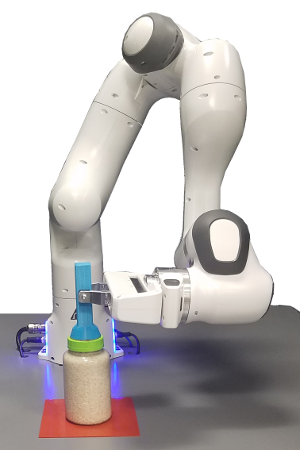}};
   \node (lsGrasp)  [left=of robotPic, yshift=-0.5cm] {\includegraphics[width=0.2\columnwidth]{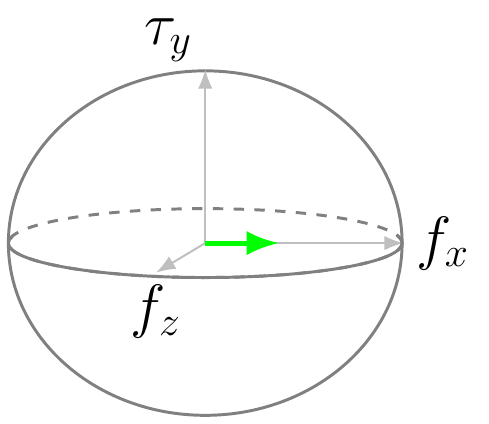}};
   \node (lsCap)    [below=of lsGrasp, yshift=1.15cm]  {\includegraphics[width=0.2\columnwidth]{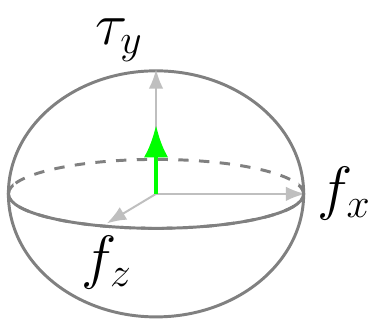}};
   \node (lsTable)  [below=of lsCap, yshift=1cm]   {\includegraphics[width=0.2\columnwidth]{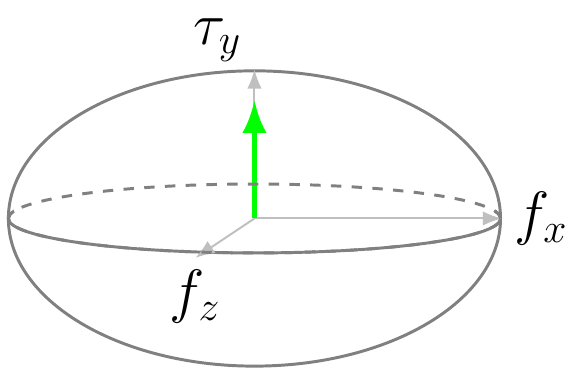}};

   \draw[gray!40, line width=0.04cm] (-5, 3) -- (-3, 3);
   \draw[gray!40, text=black, line width=0.04cm] (-5, 2.8) --+(0, 0.4) node[above] {$\tau_{min}$};
   \draw[gray!40, text=black, line width=0.04cm] (-3, 2.8) --+(0, 0.4) node[above] {$\tau_{max}$};
   \draw[gray!90, line width=0.04cm] (-4.25, 2.8) --+(0, 0.4);

   \draw[dotted, orange!50, line width=0.04cm] (-3, 3) -- (1, 2.5);
   \draw[line width=0.04cm] (1, 2.15) --+(0, 0.5);
   \draw[-{Latex[length=4pt]}, line width=0.04cm, rotate=0] (0.95, 2.5) arc [start angle=-260, end angle=90, x radius=0.4cm, y radius=0.1cm];

   \draw[gray!40, line width=0.04cm] (-5.2, 1.8) -- (-2.8, 1.8);
   \draw[gray!40, text=black, line width=0.04cm] (-5.2, 1.6) --+(0, 0.4) node[above] {$\tau_{min}$};
   \draw[gray!40, text=black, line width=0.04cm] (-2.8, 1.6) --+(0, 0.4) node[above] {$\tau_{max}$};
   \draw[gray!90, line width=0.04cm] (-4.5, 1.6) --+ (0, 0.4);

   \draw[dotted, orange!50, line width=0.04cm] (-2.8, 1.8) -- (-0.8, 2.05);
   \draw[line width=0.04cm] (-0.55, 1.7) --+(0.2, 0.5);
   \draw[-{Latex[length=4pt]}, line width=0.04cm, rotate=-25] (-1.3, 1.63) arc [start angle=-260, end angle=90, x radius=0.4cm, y radius=0.1cm];

   \draw[gray!40, line width=0.04cm] (-5.7, 0.6) -- (-2.3, 0.6);
   \draw[gray!40, text=black, line width=0.04cm] (-5.7, 0.4) --+(0, 0.4) node[above] {$\tau_{min}$};
   \draw[gray!40, text=black, line width=0.04cm] (-2.3, 0.4) --+(0, 0.4) node[above] {$\tau_{max}$};
   \draw[gray!90, line width=0.04cm] (-3.4, 0.4) --+ (0, 0.4);

   \draw[dotted, orange!50, line width=0.04cm] (-2.3, 0.6) -- (-1.5, 0.3);
   \draw[line width=0.04cm] (-1.25, 0.27) --+(0.04, 0.04);
   \draw[-{Latex[length=5pt]}, line width=0.04cm, rotate=-25] (-1.3, -0.0) arc [start angle=-260, end angle=90, x radius=0.275cm, y radius=0.275cm];

   \draw[-latex, line width=0.04cm] (-1.1, -0.9) -- (-1.1, -1.8);
   \draw[-{Latex[length=4pt]}, line width=0.04cm, rotate=0] (-0.82, -1.82) arc [start angle=50, end angle=-250, x radius=0.45cm, y radius=0.2cm];
   \draw[line width=0.04cm] (-1.1, -2) --+(0.04, 0.04);
   \draw[-{Latex[length=4pt]}, line width=0.04cm, rotate=0] (-0.78, -3.15) arc [start angle=50, end angle=-250, x radius=0.45cm, y radius=0.2cm];
   \draw[line width=0.04cm] (-1.1, -3.4) --+(0.04, 0.04);

   \draw[dotted, orange!50, line width=0.04cm] (lsGrasp.east) -- (-1.2, -1.3); 
   \draw[dotted, orange!50, line width=0.04cm] (lsCap.east) -- (-1.4, -2.1);
   \draw[dotted, orange!50, line width=0.04cm] (lsTable.east) -- (-1.4, -3.4);
 \end{tikzpicture}
\caption{Along each joint of the forceful kinematic chain, we first project the expected wrench into the subspace defined by each joint and then verify if the joint is stable under that wrench. The figure illustrate the wrench limits for each joint: For circular patch contacts, we check the friction force against a limit surface ellipsoidal model and for each robot joint we check against the 1D torque limits. }
\label{fig:wrenchSpaces}
\end{figure}

\section{Forceful Kinematic Chain}
\label{sec:kinematicChain}

We have defined a \textit{forceful kinematic chain} as the series of joints in a system, including robotic and frictional joints. 
In forceful manipulation tasks, the robot aims to exert wrenches through this chain. 
The forceful kinematic chain constraint evaluates whether each joint in the system is in equilibrium in the face of the task wrench and gravity (\figref{fig:wrenchSpaces}). 
For each joint, we propagate these wrenches to the frame of the joint and evaluate if that wrench lies in the set of wrenches that the joint can resist. 
If every joint in the chain is stable, the constraint is satisfied. 

For joint types, we consider planar frictional joints and robot manipulator joints and define the mathematical models that characterize the set of wrenches each joint type can resist.
While our treatment in this paper is limited to these joint types, alternative joints or models could easily be integrated, such as non-planar frictional joints~\citep{xu20206dfc}.

In defining planar frictional joints we consider an example joint: the robot's grasp on the blue pusher tool in \figref{fig:wrenchSpaces}. 
In the three directions of motion outside the plane of the grasp, the motion of the tool in-hand is prevented by the geometry of the hand, i.e. we assume the fingers are rigid such that the tool cannot translate or rotate by penetrating into the hand. 
Thus, any wrenches exerted in those directions are resisted kinematically by non-penetration reaction forces, which we assume are unlimited.
In the other three directions, motion within the plane of the grasp is resisted by frictional forces.
We represent the boundary of the set of possible frictional wrenches in the three dimensional friction subspace of the plane of contact with a limit surface~\citep{goyal1991planar}.
We utilize two ways to approximate the limit surface, depending on the characteristics of the planar joint (\sref{sec:ellipsoid}, \sref{sec:generalizedCone}).

For the robot's joints, the set of wrenches that can be transmitted are bound by the joint torque limits (\sref{sec:torqueLimit}).

\subsection{Limit Surface for Small Circular Patch Contacts}
\label{sec:ellipsoid}

For small circular patch contacts with uniform pressure distributions, we use an \textit{ellipsoidal approximation} of the limit surface~\citep{xydas1999modeling}.
The ellipsoid is centered in the contact frame, $w = [f_{x}, f_{z}, \tau_{y}]$, and, for isotropic friction, is defined by $w^{T}Aw~=~1$ where:
\[A~=~ 
\begin{bmatrix}
    \frac{1}{(N\mu)^2} &                    & 0  \\
                       & \frac{1}{(N\mu)^2} &   \\
    0                  &                    & \frac{1}{(Nk\mu)^2} 
\end{bmatrix}
\]
where $\mu$ is the friction coefficient, $N$ is the normal force and, for a circular patch contact we approximate $k \approx 0.6r$ where $r$ is the radius of the contact~\citep{xydas1999modeling,shi2017dynamic}.

Having transformed the wrench into the contact frame, we check if this wrench lies in the ellipsoid, which would indicate a stable contact:
\begin{equation}
\frac{f_{x}^{2}}{(N\mu)^{2}} + \frac{f_{z}^{2}}{(N\mu)^{2}} + \frac{m_{y}^{2}}{(Nk\mu)^{2}} < 1.
\label{eqn:limit_surface}
\end{equation}

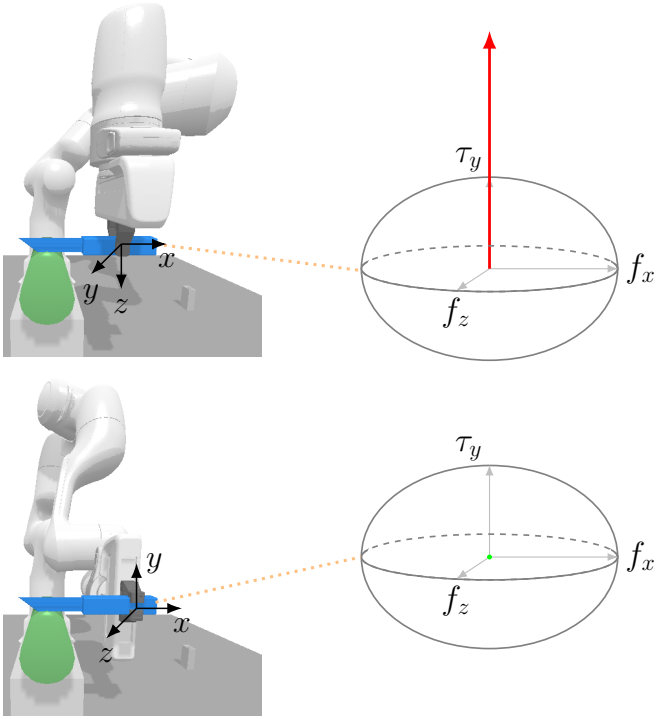
\begin{figure}[t]
\centering
 \begin{tikzpicture}[box/.style={rectangle, align=center}]

   \node (graspUnstable)                      {\includegraphics[width=0.4\columnwidth]{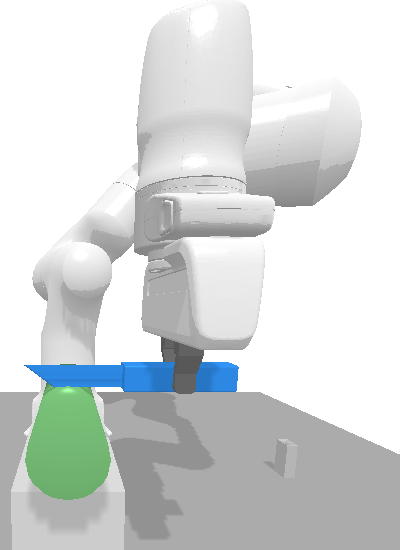}};
   \node (lsUnstable)  [right=of graspUnstable, yshift=-0.2cm] {\includegraphics[width=0.48\columnwidth]{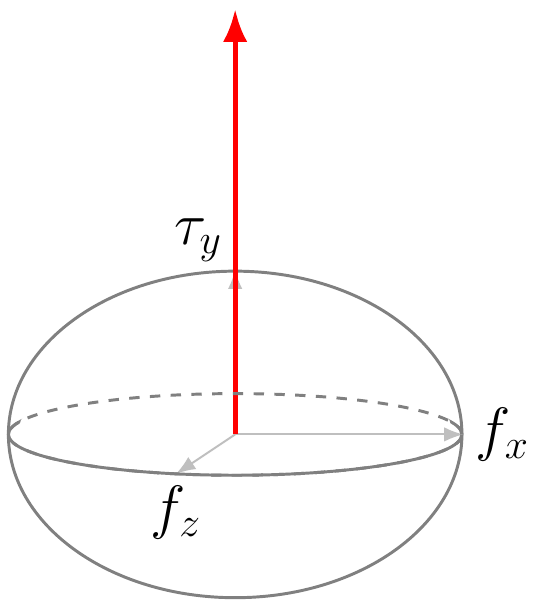}};
   \node (graspStable) [below=of graspUnstable, yshift=1.15cm]  {\includegraphics[width=0.4\columnwidth]{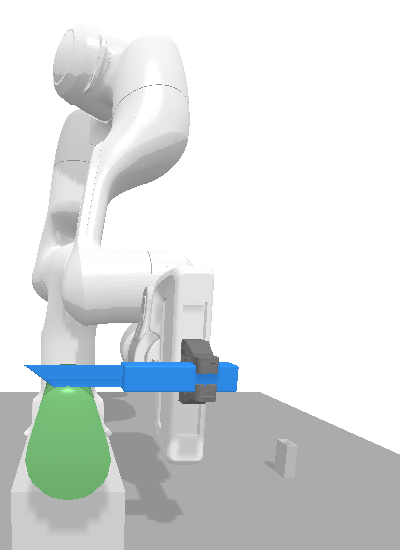}};
   \node (lsStable)    [right=of graspStable, yshift=0cm]   {\includegraphics[width=0.48\columnwidth]{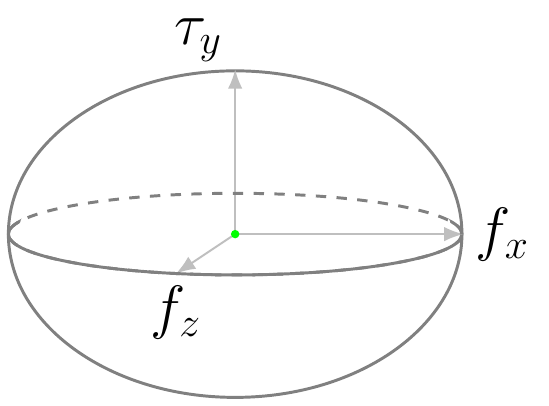}};

   \draw[dotted, orange!50, line width=0.04cm] (0.3, -0.85) -- (3, -1.2); 
   \draw[dotted, orange!50, line width=0.04cm] (0.3, -5.6) -- (3, -5.); 

   \draw[-Latex, line width=0.02cm] (-0.15, -0.85) --+ (0, -0.6) node[below] {{\large $z$}};
   \draw[-Latex, line width=0.02cm] (-0.15, -0.85) --+ (0.6, 0) node[below] {{\large $x$}};
   \draw[-Latex, line width=0.02cm] (-0.15, -0.85) --+ (-0.4, -0.4) node[below] {{\large $y$}};

   \draw[-Latex, line width=0.02cm] (0.05, -5.68) --+ (0, 0.6) node[right] {{\large $y$}};
   \draw[-Latex, line width=0.02cm] (0.05, -5.68) --+ (0.6, 0) node[below] {{\large $x$}};
   \draw[-Latex, line width=0.02cm] (0.05, -5.68) --+ (-0.4, -0.4) node[below] {{\large $z$}}; 
 \end{tikzpicture}
\caption{Here we show two possible grasps on the knife as it cuts a vegetable. For each grasp we visualize the corresponding limit surface with the propagated task wrench. 
The top grasp is not stable, as the wrench lies outside the boundary of the limit surface. 
In contrast, the bottom grasp, which leverages kinematics to resist the large torque, is stable.}
\label{fig:graspCompareExamples}
\end{figure}

As an example, in \figref{fig:graspCompareExamples} we compare two possible grasps on a knife. 
For each grasp, we visualize the limit surface and an exerted wrench, transformed to the contact frame. 
In this example we consider the first step of vegetable cutting: exerting the downward force. 


For both grasps, the friction coefficient, normal grasping force and radius of contact ($\mu,~N,~r$ respectively) are the same, so the shape of the ellipsoid is the same. 
Each grasp, however, varies the contact frame and grasp plane that the wrench is transformed into. 
In addition to impacting the magnitude of the generated torque, this defines which three directions must be resisted due to friction forces, as captured by the limit surface. 

In both grasps, transforming the downward force of this cutting action into the contact frame generates a substantial amount of torque that the grasp must resist. 

However, the grasp shown in \figref{fig:graspCompareExamples}-top, which grasps the top of the handle, largely relies on frictional forces to resist this torque. 
We can imagine if the robot were to use this grasp, the knife could pivot in the robot's hand as it moved down to cut the vegetable. 
As illustrated by the projected wrench (in red) falling outside of the ellipsoid, this grasp is unstable with respect to the forceful operation.

In contrast, in the grasp shown in \figref{fig:graspCompareExamples}-bottom, which grasps the side of the handle, the large force and torque are largely resisted kinematically and the grasp is very stable. 
Again, looking at the grasp, the geometry of the fingers, rather than friction, prevents the knife from sliding or pivoting in the hand. 

\subsection{Limit Surface for More General Patch Contacts}
\label{sec:generalizedCone}

For contacts with more irregular shapes than a circle and with less uniform pressure distributions, we directly model the contact patch as a set of point contacts, each with its own normal force (localized pressure) and its own friction limits. 
Given a contact patch we model the force it can transmit as the convex hull of generalized friction cones placed at the corners of the patch.
Generalized friction cones, based on the Coulomb friction model, represent the frictional wrench that a point contact can offer~\citep{erdmann1994representation}. 
We represent the friction cone, FC, at each point contact with a polyhedral approximation of generators:
\begin{equation} 
FC = \{ (\mu, 0, 1), (-\mu, 0, 1), (0, \mu, 1), (0, -\mu, 1) \}
\end{equation}
for a friction coefficient, $\mu$~\citep{lynch2017modern}. 
These generators can be scaled by the applied normal force. 
Given this approximation, the generalized friction cone can be written as: 
\begin{equation}
V = \{ v = J_{f}^{T} F~|~F \in {FC} \}
\label{eqn:generalized_friction_cone}
\end{equation}  
where $J_{f}^{T}$ is the Jacobian that maps contact forces $f$ from the contact frame, where FC is defined, to the object frame. 
If the exerted wrench, in the reference frame of the patch contact, lies in the convex hull of $V$, the frictional wrench can resist the exerted wrench and the contact is stable.

As an example, in the nut-twisting domain (\figref{fig:nutgrid}), we use the generalized friction cone to model the contact patch between the table and the beam holding the bolt, placing friction cones at the four corners of the beam.  
In evaluating the stability of fixturing the beam to the table via a heavy weight, the applied normal force, determined by the mass and location of the weight, is modeled as a simply supported 1D beam with a partially distributed uniform load.

\subsection{Torque Limits} 
\label{sec:torqueLimit}

The last type of joint we consider are the joints of the robot, where the limit of each joint is expressed via its torque limits. 
We relate the wrenches at the end effector to robot joint torques through the manipulator Jacobian, $J_{m}$.
Specifically, given a joint configuration $q$ and wrench $w$, the torque $\tau$ experienced at the joints is modeled by $\tau = J_{m}^{T}(q)w$.
The forceful kinematic chain is stable if the expected vector of torques $\tau$ does not exceed the robot's torque limits $\tau_{lim}$: 
\begin{equation}
J_{m}^{T}(q)w_{ext} < \tau_{lim}.
\label{eqn:torque_limit}
\end{equation}

\section{PDDLStream}
\label{sec:pddlstream}

Task and motion planning (TAMP) algorithms solve for a sequence of parameterized actions for the robot to take, also called the strategy or plan skeleton, and the hybrid parameters of those actions~\citep{garrett2020integrated}.
The parameters are discrete and continuous values such as robots, robot configurations, objects, object poses, grasping poses, regions, robot paths, wrenches, etc. 
These parameters are subject to constraints, such as requiring that all paths are collision-free. 
The parameters of forceful manipulation tasks are also subject to the forceful kinematic chain constraints, which evaluates if each joint in the chain is stable in the face of an exerted wrench. 

In order to find sequences of parameterized actions that satisfy a wide-range of constraints, we use PDDLStream, a publicly available TAMP framework~\citep{garrett2020pddlstream}. 
It has been demonstrated in a variety of robotics domains, including pick-and-place in observed and partially-observed settings~\citep{LIS267}.

In this section, we begin with some introduction to PDDL and then discuss how PDDLStream extends PDDL to enable planning over discrete and continuous parameters.

\subsection{PDDL Background}

A key challenge in solving TAMP problems is solving for the hybrid (discrete/continuous), constrained parameters. 
If all of the parameters were discrete, we could apply domain-independent classic planning algorithms from AI planning to search for sequences of actions. 
These planners use predicate language, specifically the Planning Domain Definition Language (PDDL)~\citep{mcdermott1998pddl}, to define the problem. 

PDDL is inspired by STRIPS (Stanford Research Institute Problem Solver), a problem domain specification language developed for Shakey, a mobile robot that traveled between rooms and manipulated blocks via pushing~\citep{nilsson1984shakey}. 
We next briefly describe PDDL in the context of a Shakey example.
In the following subsection we will discuss how PDDL can be augmented to address problems with hybrid parameters. 

In PDDL the state of the world is defined by a set of facts.
A fact captures a relationship among state variables. 
We denote variables with italics symbols, e.g. \variable{d}, \variable{robot}, \variable{r}. 
When defining a domain in PDDL, we specify fact types, such as \texttt{(Door \variable{d})}, \texttt{(Room \variable{r})}, \texttt{(InRoom \variable{robot} \variable{r})}. 

We denote constant variables with bold symbols, e.g. \constant{d_{1}}, \constant{robot_{shakey}}, \constant{r_{6}}. 
Hence, the fact \texttt{(Door \constant{d_{1}})} captures that in the world, there exists a door \constant{d_{1}}.
The fact \texttt{(InRoom \constant{robot_{shakey}} \constant{r_{6}})} captures that the robot \constant{robot_{shakey}} is in room \constant{r_{6}}. 

\texttt{(Door \constant{d_{1}})} is an example of a static fact, which will remain constant throughout the problem.
\texttt{(InRoom \constant{robot_{shakey}} \constant{r_{6}})} is an example of a dynamic fact, also known as a fluent, whose truth value may change over time, \textit{e.g.} as the robot moves between rooms. 

The action space is defined via a set of \textit{operators}.
Each operator is composed of a controller, a set of variables, a set of preconditions and a set of effects. 
The controller is a policy that issues a sequence of parameterized robot commands. 
Preconditions define the requirements for executing the controller, i.e. the facts, involving the operator variables, which must be true in the state in order to execute the controller. 
The effects capture how the world changes as a result of executing the controller, i.e. what facts have been added or removed from the state. 

As an example, the \texttt{go\_thru} operator is parameterized by a robot \variable{robot}, a door \variable{d}, a start room \variable{r_{s}} and a goal room \variable{r_{g}}~\citep{nilsson1984shakey}.
The controller drives the robot by following a set of waypoints, going from the start to the goal room, moving through the door. 
Two preconditions are that the robot is in room \variable{r_{s}}, i.e. \texttt{(InRoom \variable{robot}~\variable{r_{s}})}, and that the door \variable{d} connects the two rooms, i.e. \texttt{(Connect \variable{d}~\variable{r_{s}}~\variable{r_{g}})}.
The two effects of executing this operator are that the robot is no longer in the starting room and is instead in the goal room, i.e. \texttt{($\neg$ (InRoom \variable{robot}~\variable{r_{s}}))} and \texttt{(InRoom \variable{robot}~\variable{r_{g}})}, respectively.

An operator is \textit{lifted} if the variables are unassigned, i.e. \texttt{go\_thru(\variable{robot},~\variable{d},~\variable{r_{s}},~\variable{r_{g}})}. 
A \textit{ground operator} is instantiated with constant values assigned to all variables, i.e. \texttt{go\_thru(\constant{robot_{shakey}},~\constant{d_{2}},~\constant{r_{4}},~\constant{r_{7}})} where \constant{robot_{shakey}},~\constant{d_{2}},~\constant{r_{4}},~\constant{r_{7}} satisfy the constraints defined by the preconditions.
Given a ground operator, we can ground the controller, in this case mapping each room and door to fixed locations.  
The controller plans and executes a path driving to the discrete locations.

Given an set of facts that define the initial state, a set of facts defining the goal condition and a set of lifted operators, the planner searches for a sequence of ground operators to achieve the goal. 
Note that this involves solving for the sequence of operators \textit{and} for valid groundings of each operator, i.e. finding parameter values of each operator that satisfy the constraints. 

PDDL can be used to describe problems in finite domains, hence all parameter values are \textit{discrete}. 
In order to use PDDL for robotics, we need to extend this language to include \textit{continuous} values.

\begin{figure}[t!]
 \begin{tikzpicture}[box/.style={rectangle, align=center}]
   \node[box, fill=searchYellow!30, text width=25mm] (search)                     {Search using PDDL Planner};
   \node[box, fill=samplerGreen!30, text width=25mm]  (sampler)  [left=of search]  {Generate Certified Facts using Samplers};
   \node[box]                                  (plan)     [right=of search] {Plan};
   \node[box, text width=43mm]                 (inputs)   [below=of search] {Domain, \hl{Initial Facts}, Goal};
   \draw[-latex] (sampler.east) -- (search.west);
   \draw[-latex] (search.east) -- (plan.west);
   \draw[-latex] (inputs.north) -- (search.south);
   \draw[-latex] (search.north) -| (0, 1) -| (sampler.north);
 \end{tikzpicture}
\caption{Algorithmic Flow of PDDLStream. The samplers are used to certify static facts. These facts, together with the domain, initial facts and goal, serve as input to a PDDL planner, which searches for a plan. If a plan cannot be found, the algorithm generates more certified facts via the samplers.} \label{fig:pddlstreamOverview}
\end{figure}
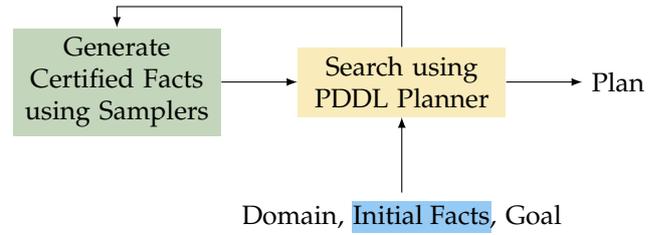

\subsection{Algorithmic Overview}

PDDLStream extends PDDL to include continuous domains by allowing for the \textit{sampling} of continuous values. 
To generate these values, PDDLStream introduces \textit{streams} that sample continuous parameters subject to constraints. 
These streams, or samplers, output static facts that the parameters are \textit{certified} to satisfy. 

For example, a configuration sampler could sample a collision-free joint configuration \constant{q} for a robot arm. 
The static fact output by this stream would be \texttt{(Conf \constant{q})}. 
A grasp sampler could generate a 6D grasp \constant{g} on an object \constant{o}, certifying the fact \texttt{(Grasp \constant{o}~\constant{g})}.
By capturing continuous parameters and their constraints using PDDL facts, PDDLStream can use discrete classic planners to plan over a hybrid search space.

To understand how sampling and search are integrated in PDDLStream it is helpful to draw an analogy to the popular motion planning algorithm of probabilistic roadmaps (PRMs)~\citep{kavraki1996probabilistic}. 
PRMs uses discrete graph search algorithms to solve a motion planning problem in the continuous space of robot configurations. 
To do this, PRMs first sample configurations and represent them as nodes in a graph. 
The edges of a PRM represent the one action the robot can make: moving from one configuration to another. 
Given an initial state, goal and graph, domain-independent graph search algorithms can be used to search for a path. 

Likewise, PDDLStream first samples parameters by executing streams to certify facts. 
Since the parameters often must satisfy a complex set of constraints, the details of the sampling procedures are often more complex than in PRMs. 
The certified facts generated by the samplers are added to the initial state. 
Given an initial state, goal and set of lifted operators, a discrete, domain-independent PDDL planner can be used to search for a plan composed of ground operator instances. 

This algorithmic procedure is illustrated in \figref{fig:pddlstreamOverview}.
The ``Incremental'' PDDLStream algorithm alternates between this process of sampling and searching until a plan is found, similarly to the way a PRM could continue to sample additional configurations and search the graph. 
In practice, we use the ``Focused'' PDDLStream algorithm, which more intelligently and efficiently samples in a lazy fashion, as detailed in~\cite{garrett2020pddlstream}.


\subsection{Specifying a Pick-and-Place Domain}
\label{sec:pddlstreamExample}
In this section we use a pick-and-place domain as an example of specifying domains using PDDLStream, highlighting how discrete and continuous parameters are captured. 
Specifying a domain requires defining a set of fact types, lifted operators and samplers. 
Both the facts and actions are specified in PDDL, while the samplers are implemented in Python. 

\begin{figure}[t!]
 \centering
 \begin{tikzpicture}[box/.style={rectangle, align=left,fill=initBlueHighlight, text width=25mm}]

   \node        (image)                        {\includegraphics[width=0.33\columnwidth]{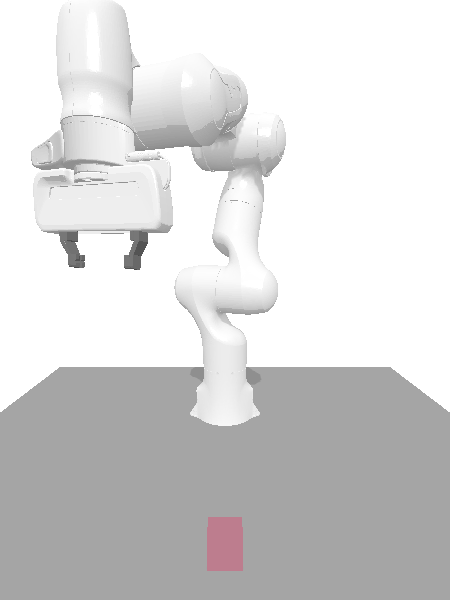}};
   \node[box]   (init)  [right=0.5cm of image] {\underline{Initial Facts:} (HandEmpty) (AtConf~\constant{q})
                                               (AtPose~\constant{o}~\constant{p}) (On \constant{o}~\constant{s})
                                               (Conf~\constant{q}) (Pose~\constant{o}~\constant{p})
                                               (Graspable~\constant{o}) (Stackable~\constant{o}~\constant{r})};
   \draw[-latex] (image.east) -- (init.west);
 \end{tikzpicture}
\caption{Pick-and-Place Example. A set of facts characterize the state. Here the robot starts at some configuration \constant{q} and a graspable pink block \constant{o} starts at some pose \constant{p} on the surface region \constant{r}.} \label{fig:pddlstreamInit}
\end{figure}

\subsubsection{Facts}~\\ Facts can capture relationships over discrete variables, like objects, and continuous variables, such as robot configurations or grasp poses.  
\figref{fig:pddlstreamInit} shows a set of facts that characterize a scene with a robot and pink block on a surface. 
The fact \texttt{(Graspable \constant{o})} defines that the object \constant{o} can be grasped, \texttt{(Pose \constant{o}~\constant{p})} defines that \constant{p} is a pose of object \constant{o}, \texttt{(AtConf \constant{q})} defines that the robot is at some configuration \constant{q}, and \texttt{(On \constant{o}~\constant{r})} defines that an object \constant{o} is on top of some surface region \constant{r}.
Here \constant{o} and \constant{r} are discrete parameters while \constant{p} and \constant{q} are continuous. 

Facts can also be derived from other facts, e.g. the fact that a robot is holding an object \variable{o}, \texttt{(Holding \variable{o})}, is true if there exists a grasp \variable{g} such that the robot is at some grasp, i.e. \texttt{(AtGrasp \variable{o}~\variable{g})}.
These are called \textit{derived facts}. 

\subsubsection{Operators}~\\ Operators are parameterized by discrete and continuous values. 
As an example, the \texttt{pick} action is parameterized by an object to be grasped \variable{o}, the pose of that object \variable{p}, a grasp on the object \variable{g}, a configuration \variable{q} and a trajectory \variable{t}.
Two preconditions of the action are that the robot isn't already holding something, \texttt{(HandEmpty)}, and that the object is at the pose \texttt{(AtPose \variable{o}~\variable{p})}. 
Two effects of the operator are that the robot's hand is no longer empty, \texttt{($\neg$ (HandEmpty))} and that the robot is now at the grasp, \texttt{(AtGrasp \variable{o}~\variable{g})}. 
We define the pick controller to be a series of commands: executing trajectory \variable{t} to move towards the object, closing the hand, executing trajectory \variable{t} in reverse to back up.


\begin{figure}[t!]
  \centering
  \begin{tikzpicture}[
     sampler/.style={rectangle, fill=samplerGreen!30, align=center},
     io/.style={rectangle, align=center},
     every node/.style={node distance=0.5cm, font=\small},]

   \node[io, text width=70mm]      (iik)                     {Input: \variable{o}, \variable{g}, \variable{p} s.t. (Grasp \variable{o}~\variable{g}) (Pose \variable{o}~\variable{p})};
   \node[sampler, text width=30mm] (ik)     [below=of iik]   {inverse-kinematics};
   \node[io, text width=80mm]      (oik)    [below=of ik]    {Output: \variable{q}, \variable{t} s.t. (Conf \variable{q}) (Traj \variable{t}) (Kin \variable{o}~\variable{p}~\variable{g}~\variable{q}~\variable{t})};
   \node[io, text width=50mm]      (itraj)  [below=of oik]   {Input: \variable{t} s.t. (Traj \variable{t})};
   \node[sampler, text width=30mm] (traj)   [below=of itraj] {check-traj-collision};
   \node[io, text width=50mm]      (otraj)  [below=of traj]  {Output s.t. (UnsafeTraj \variable{t})}; 

   \node[io, text width=35mm]      (ograsp) [above left=1cm of iik,xshift=4cm] {Output: \variable{g} s.t. (Grasp \variable{o}~\variable{g})};
   \node[sampler, text width=22mm] (grasp)  [above=of ograsp]   {sample-grasp};
   \node[io, text width=35mm]      (igrasp) [above=of grasp]    {Input: \variable{o} s.t. (Graspable \variable{o})};

   \node[io, text width=45mm]      (opose)  [above right=1cm of iik,xshift=-4cm] {Output: \variable{p} s.t. (Pose~\variable{o}~\variable{p}) (Supported \variable{o}~\variable{p}~\variable{r})};
   \node[sampler, text width=22mm] (pose)   [above=of opose]     {sample-pose};
   \node[io, text width=35mm]      (ipose)  [above=of pose]      {Input: \variable{o}, \variable{r} s.t. (Stackable \variable{o}~\variable{r})};

   \draw[-latex] (ipose.south) -- (pose.north);
   \draw[-latex] (pose.south) -- (opose.north);
   \draw[dotted,-latex] (opose.south) -- (iik.north);
   \draw[-latex] (igrasp.south) -- (grasp.north);
   \draw[-latex] (grasp.south) -- (ograsp.north);
   \draw[dotted,-latex] (ograsp.south) -- (iik.north);
   \draw[-latex] (iik.south) -- (ik.north);
   \draw[-latex] (ik.south) -- (oik.north);
   \draw[dotted,-latex] (oik.south) -- (itraj.north);
   \draw[-latex] (itraj.south) -- (traj.north);
   \draw[-latex] (traj.south) -- (otraj.north);
  \end{tikzpicture}
 
  \caption{Pick-and-Place Example. Each sampler takes as input some values such that those values satisfy some constraint. The samplers (highlighted in green), output either new parameter values that are certified to satisfy some constraint or simply a certification that the inputs satisfy a constraint. Samplers can be conditioned upon each other such that the output of one sampler is the input to another, as shown by the dotted line.} \label{fig:pddlstreamStreams}
\end{figure}
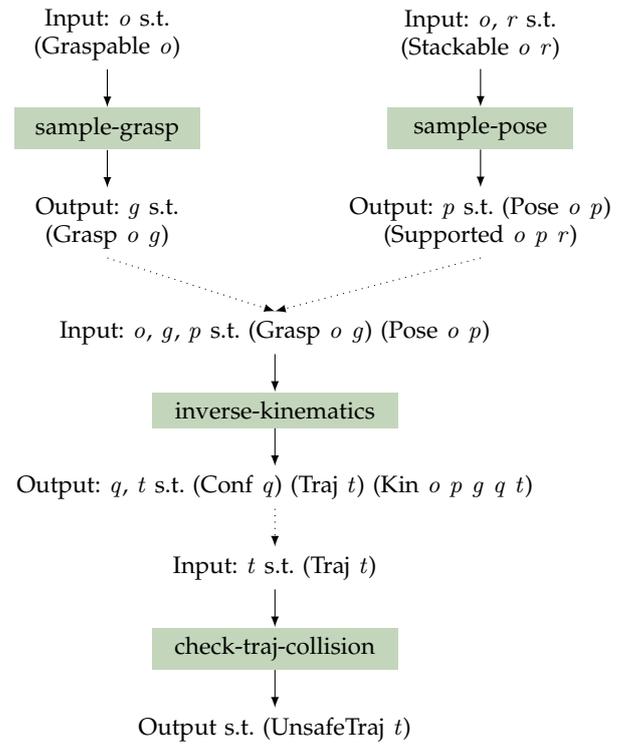

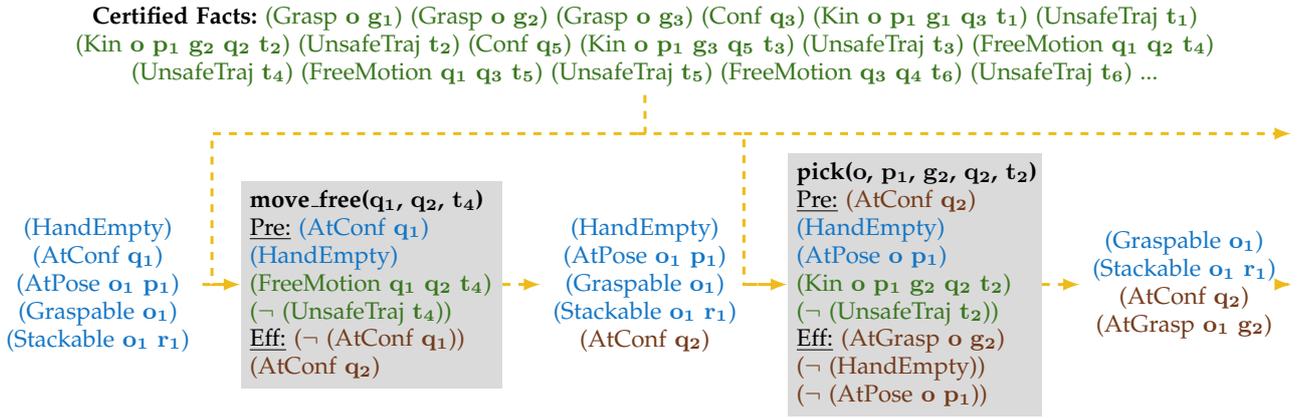
\begin{figure*}[t!]
  \centering
  \begin{tikzpicture}[
     action/.style={rectangle, fill=actionGrey!90!black, align=left},
     facts/.style={rectangle, align=center},
     arrow/.style={dashed, searchYellow, line width=0.04cm},
     every node/.style={node distance=0.5cm, font=\small}]

   \node[facts, text width=25mm]  (facts0)                    {\textcolor{initBlue}{(HandEmpty) (AtConf \constant{q_{1}}) (AtPose \constant{o_{1}}~\constant{p_{1}}) 
                                                                 (Graspable \constant{o_{1}}) (Stackable \constant{o_{1}}~\constant{r_{1}})}};
   \node[action, text width=32mm] (move)    [right=of facts0] {\textbf{move\_free(\constant{q_{1}},~\constant{q_{2}},~\constant{t_{4}})}
                                                                \underline{Pre:} \textcolor{initBlue}{(AtConf \constant{q_{1}}) (HandEmpty)}
                                                                     \textcolor{samplerGreen}{(FreeMotion \constant{q_{1}}~\constant{q_{2}}~\constant{t_{4}}) 
                                                                     ($\neg$~(UnsafeTraj \constant{t_{4}}))}
                                                                \underline{Eff:} \textcolor{effectBrown}{($\neg$ (AtConf \constant{q_{1}}))
                                                                     (AtConf \constant{q_{2}})}};
   \node[facts, text width=25mm]  (facts1)  [right=of move]   {\textcolor{initBlue}{(HandEmpty) (AtPose \constant{o_{1}}~\constant{p_{1}}) 
                                                                 (Graspable \constant{o_{1}}) (Stackable \constant{o_{1}}~\constant{r_{1}})} 
                                                                \textcolor{effectBrown}{(AtConf \constant{q_{2}})}};
   \node[action, text width=31mm] (pick)    [right=of facts1] {\textbf{pick(\constant{o},~\constant{p_{1}},~\constant{g_{2}},~\constant{q_{2}},~\constant{t_{2}})}
                                                                \underline{Pre:} \textcolor{effectBrown}{(AtConf \constant{q_{2}})} 
                                                                                 \textcolor{initBlue}{(HandEmpty) 
                                                                                 (AtPose \constant{o}~\constant{p_{1}})}
                                                                                 \textcolor{samplerGreen}{(Kin~\constant{o}~\constant{p_{1}}~\constant{g_{2}}~\constant{q_{2}}~\constant{t_{2}}) 
                                                                                 ($\neg$~(UnsafeTraj~\constant{t_{2}}))}
                                                                \underline{Eff:} \textcolor{effectBrown}{(AtGrasp \constant{o}~\constant{g_{2}}) 
                                                                                 ($\neg$~(HandEmpty)) 
                                                                                 ($\neg$~(AtPose \constant{o}~\constant{p_{1}}))}};
   \node[facts, text width=25mm]  (facts2)  [right=of pick]   {\textcolor{initBlue}{(Graspable \constant{o_{1}}) (Stackable \constant{o_{1}}~\constant{r_{1}})}
                                                                 \textcolor{effectBrown}{(AtConf \constant{q_{2}}) (AtGrasp \constant{o_{1}}~\constant{g_{2}})}};

   \node[facts, text width=\textwidth] (cert) [above=1.5cm of facts1] {\textbf{Certified Facts:} 
                                                               \textcolor{samplerGreen}{(Grasp \constant{o}~\constant{g_{1}})  
                                                                (Grasp~\constant{o}~\constant{g_{2}}) 
                                                                (Grasp~\constant{o}~\constant{g_{3}}) (Conf~\constant{q_{3}}) 
                                                                (Kin~\constant{o}~\constant{p_{1}}~\constant{g_{1}}~\constant{q_{3}}~\constant{t_{1}}) 
                                                                (UnsafeTraj~\constant{t_{1}}) 
                                                                (Kin~\constant{o}~\constant{p_{1}}~\constant{g_{2}}~\constant{q_{2}}~\constant{t_{2}}) 
                                                                (UnsafeTraj~\constant{t_{2}}) 
                                                                (Conf~\constant{q_{5}}) 
                                                                (Kin~\constant{o}~\constant{p_{1}}~\constant{g_{3}}~\constant{q_{5}}~\constant{t_{3}})  
                                                                (UnsafeTraj~\constant{t_{3}}) 
                                                                (FreeMotion~\constant{q_{1}}~\constant{q_{2}}~\constant{t_{4}}) 
                                                                (UnsafeTraj~\constant{t_{4}}) 
                                                                (FreeMotion~\constant{q_{1}}~\constant{q_{3}}~\constant{t_{5}}) 
                                                                (UnsafeTraj~\constant{t_{5}})
                                                                (FreeMotion~\constant{q_{3}}~\constant{q_{4}}~\constant{t_{6}}) 
                                                                (UnsafeTraj~\constant{t_{6}}) ...}};

   \draw[arrow,-latex] (facts0.east) -- (move.west);
   \draw[arrow,-latex] (move.east) -- (facts1.west);
   \draw[arrow,-latex] (facts1.east) -- (pick.west);
   \draw[arrow,-latex] (pick.east) -- (facts2.west);
   \draw[arrow] (cert.south) -- (7.2,2);
   \draw[arrow] (7.1, 2)  -| (1.5, 0) -| (move.west);
   \draw[arrow] (7.1, 2)  -| (8.5, 0) -| (pick.west);
   \draw[arrow,-latex] (8.5, 2) -- (15.7, 2);
   \draw[arrow,-latex] (facts2.east) -- (15.7, 0);
  \end{tikzpicture}

\caption{Pick-and-Place Example. Expanded view of the PDDLStream search procedure. The state is composed of facts from the initial set of facts (in blue) and the set of certified static facts generated by the samplers (in green). An action is feasible if all of the facts of the preconditions are met. The state is then updated with the resulting effects (in brown). In this example, the first action is to \texttt{move} from configuration \constant{q_{1}} to \constant{q_{2}}. Having taken this action, the next is to pick up an object \constant{o} at pose \constant{p_{1}} using grasp \constant{g_{2}}. The result of the search would be the sequence of ground operators, i.e. [move\_free(\constant{q_{1}},~\constant{q_{2}},~\constant{t_{4}}), pick(\constant{o},~\constant{p_{1}}, \constant{g_{2}}, \constant{q_{2}}, \constant{t_{2}})].} \label{fig:pddlstreamSearch}
\end{figure*}

\subsubsection{Samplers}~\\
Samplers are conditional generators that output static facts that \textit{certify} that the set of parameters satisfy some set of constraints. 
\figref{fig:pddlstreamStreams} shows several possible samplers, and how samplers can be conditioned on the output of other samplers.

For example, \texttt{sample-grasp} takes in an object \variable{o} that is constrained to be graspable, and returns a grasp \variable{g} that is certified to be a grasp of the object, as represented by the generation of the fact \texttt{(Grasp \variable{o}~\variable{g})}.
Likewise, \texttt{sample-pose} takes in an object \variable{o} and region \variable{r} such that the object can be stacked on the region, i.e. \texttt{(Stackable \variable{o}~\variable{r})}.
The output of this sampler is a pose \variable{p} that is certified to be a valid pose of the object \variable{o} and that if object \variable{o} is at pose \variable{p}, that object \variable{o} is supported by region $r$.

The grasp \variable{g} and pose \variable{p} generated by these two samplers, along with the object \variable{o}, can be inputs to an \texttt{inverse-kinematics} sampler which outputs a configuration \variable{q} and trajectory \variable{t}.
The relationship between the variables is captured in the static fact \texttt{(Kin \variable{o}~\variable{p}~\variable{g}~\variable{q}~\variable{t})}: the trajectory \variable{t} starting from configuration \variable{q} grasps object \variable{o} at pose \variable{p} with grasp \variable{g}.

In addition to generative samplers, there are also test samplers which evaluate to true or false based on whether the input variables satisfy some constraint.
These samplers do not generate new parameters and instead only add facts about existing parameters. 

As an example, a test stream is used evaluate whether a trajectory \variable{t} is collision-free. 
This is a condition over the entire state, i.e. the trajectory must be collision-free with respect to all objects in the environment. 
Evaluating if a trajectory is collision-free by an universal qualifier (\texttt{forall}) is expensive for PDDL planners. 
Therefore, instead we can use negation to evaluate an existential quantifier (\texttt{exists}). 
Specifically, we use the test sampler \texttt{check-traj-collision} to certify the fact \texttt{(UnsafeTraj \variable{t})} if there exists any object \variable{o} at pose \variable{p} such that trajectory \variable{t} is in collision with \variable{o} at \variable{p}\footnote{Both \variable{o} and \variable{p} are omitted from \figref{fig:pddlstreamStreams} for clarity.}. 
We can then enforce that trajectories are collision-free by using \texttt{(not (Unsafe \variable{t}))} as a precondition. 


\subsection{Search Procedure}


Given the domain specified in the previous subsection, we consider one step of the search, visualized in \figref{fig:pddlstreamSearch}

The state is defined by all of the facts that are true. 
At the beginning of the search, this is the set of initial facts (in blue) and the set of facts certified by the samplers (in green).
In this example pick-and-place domain, \figref{fig:pddlstreamInit} gives the initial facts and \figref{fig:pddlstreamStreams} visualizes some of the streams.

As stated above, the facts certified by the samplers are static. 
As an example, if a grasp sampler generates a grasp on an object, the fact capturing this grasp, \texttt{(Grasp \constant{o}~\constant{g})}, is always true.
Thus, in the search, the certified facts output by the samplers are always a part of the state.
The facts from the initial state may be static facts, such as \texttt{(Graspable \constant{o})}, or fluents, such as \texttt{(AtConf \constant{q_{1}})}.
These latter types of facts may be added or removed, as a result of the operators' effects, as the robot acts in the environment. 

Given the state composed of all of these facts, the planner now conducts a search for a sequence of actions that can be taken to achieve the goal.
For an action to be valid, all of its preconditions must be met, i.e. all of the precondition facts must be true in the state. 
The facts could be true either because the fact was certified by a sampler, or because it was part of the initial state or because it was added to the state as the effect of a previous action.

In \figref{fig:pddlstreamSearch}, the \texttt{move\_free} action can be taken because all of the preconditions are satisfied. 
If the planner elected to take this action, the effects of the action would update the current state (in brown).
The search continues forward based on this updated state, searching for valid actions.

Following the algorithmic loop in \figref{fig:pddlstreamOverview}, if the search is unsuccessful in finding a sequence of actions to the goal, the algorithm would generate more certified facts via the samplers and search again.

\begin{table*}[t!]
\begin{subtable}{\textwidth}
 \centering
 \begin{tabular}{l c c}
   \hline
   Operator & Preconditions & Effects \\ [0.5ex]
   \hline\hline
   \rowcolor{addedKinematic}
   move\_free & 
   \makecell[cc]{(AtConf $a~q_{1}$) $\wedge$ (HandEmpty $a$) $\wedge$ \\
                 \textcolor{samplerGreen}{(FreeMotion $a~q_{1}~q_{2}~t$)} $\wedge$ ($\neg$ (TrajUnsafe $a~t$))} & 
   \makecell[cc]{($\neg$ (AtConf $a~q_{1}$)) \\    
                 $\wedge$ (AtConf $a~q_{2}$)} \\
   \hline
   \rowcolor{addedKinematic}
   move\_holding & 
   \makecell[cc]{(AtConf $a~q_{1}$) $\wedge$ (Movable $o$) $\wedge$ (AtGrasp $a~o~g_{o}$) $\wedge$  \\
                 \textcolor{samplerGreen}{(HoldingMotion $a~o~g_{o}~q_{1}~q_{2}~t$)} $\wedge$ 
                 ($\neg$ (TrajUnsafe $a~t$))} & 
   \makecell[cc]{($\neg$ (AtConf $a~q_{1}$)) \\    
                 $\wedge$ (AtConf $a~q_{2}$)} \\
   \hline
   \rowcolor{addedKinematic}
   pick & 
   \makecell[cc]{(AtConf $a~q$) $\wedge$ (HandEmpty $a$) $\wedge$ ((Movable $o$) $\wedge$  \\ (AtPose $o~p_{o}$) $\wedge$
                 \textcolor{samplerGreen}{(Kin $a~o~p_{o}~g_{o}~q~t$)} $\wedge$ ($\neg$ (TrajUnsafe $a~t$))} & 
   \makecell[ccc]{(AtGrasp $a~o~g_{o}$) \\ 
                 $\wedge$ ($\neg$ (AtPose $o~p_{o}$)) \\ 
                 $\wedge$ ($\neg$ (HandEmpty $a$))} \\
   \hline
   \rowcolor{addedKinematic}
   place &
   \makecell[cc]{(AtConf $a~q$) $\wedge$ (AtGrasp $a~o~g_{o}$) $\wedge$  \\
                 \textcolor{samplerGreen}{(Kin $a~o~p_{o}~g_{o}~q~t$)} $\wedge$ ($\neg$ (TrajUnsafe $a~t$))} & 
   \makecell[ccc]{(AtPose $o~p_{o}$) \\
                 $\wedge$ (HandEmpty $a$) \\ 
                 $\wedge$ ($\neg$ (AtGrasp $a~o~g_{o}$))} \\
   \hline
   \rowcolor{addedTask}
   hand\_twist & 
   \makecell[ccc]{(AtConf $a~q_{0}$) $\wedge$ (HandEmpty $a$) $\wedge$ (Nut $o_{n}$) $\wedge$ (Bolt $o_{b}$) $\wedge$ \\
                 (AtPose $o_{n}~p_{n}$) $\wedge$ (On $o_{n}~o_{b}$) $\wedge$  
                 (Fixtured $o_{b}~w$) $\wedge$ \\
                 \textcolor{samplerGreen}{(StableGrasp $o_{n}~g_{n}~w$)} $\wedge$ 
                 \textcolor{samplerGreen}{(StableJoints $a~t~w$)} $\wedge$ \\ 
                 \textcolor{samplerGreen}{(NutHandMotion $a~o_{n}~p_{n}~g_{n}~w~q_{0}~q_{1}~t$)}) $\wedge$ ($\neg$ (TrajUnsafe $a~t$))} & 
   \makecell[ccc]{(Twisted $o_{n}~w$) \\
                 $\wedge$ ($\neg$(AtConf $a~q_{0}$)) \\ 
                 $\wedge$ (AtConf $a~q_{1}$))} \\
   \hline
   \rowcolor{addedTask}
   tool\_twist &
   \makecell[ccc]{ (AtConf $a~q_{0}$) $\wedge$ (Nut $o_{n}$) $\wedge$ (Bolt $o_{b}$) $\wedge$ 
                 (AtPose $o_{n}~p_{n}$) $\wedge$ (On $o_{n}~o_{b}$) $\wedge$    \\ 
                 (Spanner $o_{t}$) $\wedge$ (AtGrasp $a~o_{t}~g_{t}$) $\wedge$  
                 (Fixtured $o_{b}~w$) $\wedge$  \\
                 \textcolor{samplerGreen}{(StableGrasp $o_{t}~g_{t}~w$)} $\wedge$ 
                 \textcolor{samplerGreen}{(StableJoints $a~t~w$)} $\wedge$ \\
                 \textcolor{samplerGreen}{(NutToolMotion $a~o_{t}~o_{n}~p_{n}~g_{t}~w~q_{0}~q_{1}~t$)}) $\wedge$ ($\neg$ (TrajUnsafe $a~t$))} & 
   \makecell[ccc]{(Twisted $o_{n}~w$) \\
                  $\wedge$ ($\neg$(AtConf $a~q_{0}$)) \\
                  $\wedge$ (AtConf $a~q_{1}$))} \\
   \hline
 \end{tabular}
\caption{Operators} \label{table:nutActions}
\end{subtable}
\vspace{0.5cm}
\begin{subtable}{\textwidth}
 \centering
 \begin{tabular}{l l}
   \hline
   Derived Facts & Definition \\ 
   \hline\hline
   \rowcolor{addedKinematic}
   (On $o~r$) & $\exists p_{o}$ (\textcolor{samplerGreen}{(Supported $o~p_{o}~r$)} $\wedge$ (AtPose $o~p_{o}$)) \\
   \hline
   \rowcolor{addedKinematic}
   (Holding $o$) & $\exists a, g_{o}$ (AtGrasp $a~o~g_{o}$) \\ 
   \hline
   \rowcolor{addedKinematic}
   (TrajUnsafe $a_{1}~t_{1}$) & 
   \makecell[ll]{($\exists o, p$ ((AtPose $o~p$) $\wedge$ ($\neg$ \textcolor{samplerGreen}{(ObjCollisionFree $a_{1}~t_{1}~o~p$)}))) $\vee$ \\ 
                 ($\exists a_{2}, q_{2}$ ((AtConf $a_{2}~q_{2}$) $\wedge$ ($\neg$ \textcolor{samplerGreen}{(ArmCollisionFree $a_{1}~t_{1}~a_{2}~q_{2}$)})))} \\
   \hline
   \rowcolor{addedFM}
   (HoldingFixtured $o~w$) & $\exists g, a$ (AtGrasp $a~o~g$) $\wedge$ \textcolor{samplerGreen}{(StableGrasp $o~g~w$)}) \\ 
   \hline 
   \rowcolor{addedFM}
   (WeightFixtured $o~w$) & $\exists o_{1}, p_{1}$ ((Weight $o_{1}$) $\wedge$ (AtPose $o_{1}~p_{1}$) $\wedge$ (On $o_{1}~o$) $\wedge$ \textcolor{samplerGreen}{(StableWeighDown $o_{1}~p_{1}~o~w$)}) \\ 
   \hline
   \rowcolor{addedFM}
   (Fixtured $o~w$) & ((HoldingFixtured $o~w$) $\vee$ (WeightFixtured $o~w$)) \\
   \hline
 \end{tabular}
\caption{Derived Facts} \label{table:nutDerived}
\end{subtable}
\vspace{0.5cm}
\begin{subtable}{\textwidth}
 \centering
 \begin{tabular}{l c c c}
   \hline
   Sampler & Inputs & Outputs & Certified Facts \\
   \hline\hline
   \rowcolor{addedKinematic}
   sample-pose & $o~r$ & $p_{o}$ & 
      \textcolor{samplerGreen}{(Pose $o~p_{o}$) $\wedge$ (Supported $o~p_{o}~r$)}\\
   \hline
   \rowcolor{addedKinematic}
   sample-grasp & $a~o$ & $g_{o}$ & 
      \textcolor{samplerGreen}{(Grasp $a~o~g_{o}$)} \\
   \hline
   \rowcolor{addedKinematic}
   inverse-kinematics & $a~o~p_{o}~g_{o}$ & $q~t$ & 
      \textcolor{samplerGreen}{(Kin $a~o~p_{o}~g_{o}~q~t$) $\wedge$ (Conf $q$) $\wedge$ (Traj $t$)} \\
   \hline
   \rowcolor{addedKinematic}
   plan-free-motion & $a~q_{0}~q_{1}$ & $t$ & 
      \textcolor{samplerGreen}{(FreeMotion $a~q_{1}~q_{2}~t$) $\wedge$ (Traj $t$)} \\
   \hline
   \rowcolor{addedKinematic}
   plan-holding-motion & $a~q_{1}~q_{2}~o~g_{o}$ & $t$ & 
     \textcolor{samplerGreen}{(HoldingMotion $a~o~g_{o}~q_{1}~q_{2}~t$) $\wedge$ (Traj $t$)} \\
   \hline
   \rowcolor{addedKinematic}
   test-arm-collision & $a_{1}~a_{2}~t_{1}~q_{2}$ & & 
      \textcolor{samplerGreen}{(ArmCollisionFree $a_{1}~a_{2}~t_{1}~q_{2}$)} \\
   \hline
   \rowcolor{addedKinematic}
   test-obj-collision & $a_{1}~t_{1}~o~p$ & & 
      \textcolor{samplerGreen}{(ObjCollisionFree $a_{1}~t_{1}~o~p$)} \\
   \hline
   \rowcolor{addedFM}
   test-grasp-stable & $a~o~w~g_{o}$ & & 
      \textcolor{samplerGreen}{(StableGrasp $o~g_{o}~w$)} \\
   \hline
   \rowcolor{addedFM}
   test-joints-stable & $a~t~w$ & & 
      \textcolor{samplerGreen}{(StableJoints $a~t~w$)} \\
   \hline
   \rowcolor{addedFM}
   test-weight-stable & $o_{1}~p_{1}~o~w$ & & 
      \textcolor{samplerGreen}{(StableWeighDown $o_{1}~p_{1}~o~w$)} \\
   \hline
   \rowcolor{addedTask}
   plan-nut-hand & $a~o_{n}~p_{n}~g_{n}~w$ & $q_{0}~q_{1}~t$ & 
      \makecell[cc]{\textcolor{samplerGreen}{(NutHandMotion $a~o_{n}~p_{n}~g_{n}~w~q_{0}~q_{1}~t$) $\wedge$ } \\
                    \textcolor{samplerGreen}{(Conf $q_{1}$) $\wedge$ (Conf $q_{2}$) $\wedge$ (Traj $t$)}} \\
   \hline
   \rowcolor{addedTask}
   plan-nut-tool & $a~o_{t}~o_{n}~p_{n}~g_{n}~w$ & $q_{0}~q_{1}~t$ & 
       \makecell[cc]{\textcolor{samplerGreen}{(NutToolMotion $a~o_{t}~o_{n}~p_{n}~g_{t}~w~q_{0}~q_{1}~t$) $\wedge$}  \\
                     \textcolor{samplerGreen}{(Conf $q_{1}$) $\wedge$ (Conf $q_{2}$) $\wedge$ (Traj $t$)}} \\
   \hline
 \end{tabular}
\caption{Samplers} \label{table:nutSamplers}
\end{subtable}
\caption{The specification of the nut twisting domain via the lifted operators, derived facts and samplers. Elements colored in light grey are common across all forceful manipulation domains. Elements colored in darker grey are specific to the nut-twisting domain. Throughout the table we use the symbols: \variable{a} is a robot arm, \variable{o} is an object, \variable{p_{o}} is a pose of object \variable{o}, \variable{g_{o}} is a grasp on object \variable{o}, \variable{q_{i}} is a configuration, \variable{r} is a region, \variable{t} is a trajectory, \variable{w} is a wrench. Specific to this domain: \variable{o_{n}}, \variable{o_{b}} and \variable{o_{t}} refer to the nut, bolt and spanner, respectively.} \label{table:nut}
\end{table*}

\section{Incoporating Force into Planning}
\label{sec:forceplanning}

PDDLStream provides a framework for solving TAMP problems. 
\sref{sec:pddlstreamExample} showed how to specify the standard pick-and-place domain, highlighting the fact types, lifted operators and samplers needed to capture the actions and the domain constraints, which relate to kinematic and geometric feasibility. 

In order to leverage PDDLStream for forceful manipulation tasks, and thus extend PDDLStream's range of applicability, we encode the the forceful kinematic chain constraint and the fixturing requirement.
We do so by adding fact types, lifted operators and samplers. 
Additionally, each domain has domain-specific elements, although we find that many samplers are reused across domains. 
As an illustrative example for how these pieces come together, in \sref{sec:pddlstreamForce} we detail the domain specification for the nut-twisting task.  
Through this example we also explore what modeling effort is required to specify a domain. 

In addition to finding plans, in \sref{sec:robust} we discuss how we use cost-sensitive planning in PDDLStream to find robust plans for forceful manipulation tasks.

\subsection{PDDLStream for Forceful Manipulation}
\label{sec:pddlstreamForce}


We use nut-twisting as an example of how the framework of PDDLStream is used for forceful manipulation tasks.
The lifted operators, derived facts and samplers for this domain are defined in \tref{table:nut}. 
We first require the standard pick-and-place operators (along with their related derived facts and samplers), some of which were already described in \sref{sec:pddlstreamExample}: \texttt{move\_free}, \texttt{move\_holding}, \texttt{pick}, \texttt{place}.

We next include the elements (in light grey) common across all of our forceful manipulation domains, which capture the forceful kinematic chain constraint and enable fixturing.
As discussed in \sref{sec:fixturing}, there are several different fixturing strategies so in this domain we focus on two of them (\texttt{HoldingFixtured} and \texttt{WeightFixtured}), while the rest are detailed in \apref{appendix:domains}.
In all forceful manipulation domains we add the fact type \texttt{(Wrench \variable{w})} where \variable{w} is a 6D wrench and a coordinate frame.

Finally, we define two domain-specific operators, \texttt{hand\_twist} and \texttt{tool\_twist} (and related samplers, all colored in darker grey), which correspond to the robot twisting the nut with its hand or with a grasped tool, respectively.

\subsubsection{Forceful Kinematic Chain Constraint}
\label{sec:fkcPDDL}

To incorporate the forceful kinematic chain constraint into the PDDLStream framework, we assess the stability of the chain using test samplers.
The facts certified by the test samplers serve as preconditions for the operators that exert forceful operations.

For example, to assess the planar frictional joints formed by the robot's grasp, we define fact \texttt{(StableGrasp \variable{o}~\variable{g}~\variable{w})} which is true if the grasp \variable{g} on object \variable{o} is stable under the wrench \variable{w}. 
This fact is certified by the test sampler \texttt{test-grasp-stable}, which uses the limit surface model discussed in \sref{sec:ellipsoid} to evaluate the stability of the joint.
 
For an operator that applies a forceful operation using a grasped object, we use \texttt{(StableGrasp \variable{o}~\variable{g}~\variable{w})} as a precondition to constrain the grasp to be stable. 
In the context of the nut-twisting domain, \texttt{hand\_twist} uses this precondition to evaluate the robot's grasp on the nut and \texttt{tool\_twist} uses it to evaluate the robot's grasp on the tool.

We also use the \texttt{(StableGrasp \variable{o}~\variable{g}~\variable{w})} fact to derive the \texttt{(HoldingFixtured \variable{o}~\variable{w})} fact, which defines if an object \variable{o} is fixtured from wrench \variable{w} using some grasp \variable{g}. 

As another example, to evaluate if robot joints are stable we define the fact \texttt{(StableJoints \variable{a}~\variable{t}~\variable{w})} which is true if the trajectory \variable{t} executed on robot arm \variable{a} is stable under wrench \variable{w}. 
This is certified by the test sampler \texttt{test-joints-stable}, which evaluates the torques experienced at each configuration in the trajectory are within the arm's torque limits.

\subsubsection{Fixturing}
\label{sec:fixturePDDL}

In addition to the forceful kinematic chain constraint, we require that while the robot is exerting a forceful operation on an object, the object must be fixtured. 
We propose implementing a variety of fixturing methods through various lifted operators and derived facts, which use forceful kinematic chain test samplers to evaluate the stability of the fixturing chain. 

In the context of the nut-twisting domain, we restrict our focus to two fixturing methods: fixturing an object by holding it or by weighing it down with another heavy object.
Thus, in this domain, the fact \texttt{(Fixtured \variable{o}~\variable{w})} is satisfied if either \texttt{(HoldingFixtured \variable{o}~\variable{w})} or \texttt{(WeightFixtured \variable{o}~\variable{w})} are true.
The latter fact is derived from the fact \texttt{(StableWeighDown \variable{o_{1}}~\variable{p_{1}}~\variable{o}~\variable{w})}, which is certified by the test sampler \texttt{test-weight-stable}.
This sampler uses the generalized friction cone from \sref{sec:generalizedCone} to evaluate stability.

Neither of these two fixturing methods required adding new operators, since sequences of moves, picks and places can be constructed to either stably grasp the fixturing object or have a robot place a weighing-down object on the fixtured object.
However, other fixturing methods (detailed in \apref{appendix:domains}), such as operating a vise, include adding operators, as well as test samplers and facts, to the domain.

\subsubsection{Domain-Specific Actions}

Given the forceful manipulation extensions, we now detail the domain-specific additions.
In the nut-twisting domain the robot can impart the forceful operation to twist the nut either by making contact with its fingers or through a grasped spanner.
To enable this, we define two new operators in \tref{table:nut}.

The controller for the \texttt{hand\_twist} operator grasps the nut with the hand, forcefully twists the nut with the hand and the release the nut. 
Likewise, the controller for \texttt{tool\_twist} makes contact with the nut via the grasped spanner, forcefully twists the nut with the spanner and then breaks contact between the nut and the spanner. 
The trajectory for each operator is generated by the samplers \texttt{plan-nut-hand} and \texttt{plan-nut-tool}, respectively. 
The twisting step of the trajectory involves planning a path that applies the forceful operation.
Across all domains, we use Cartesian impedance control to exert wrenches, as detailed in \apref{appendix:controller}.

As stated in \sref{sec:fkcPDDL}, the forceful kinematic chain constraint is implemented as preconditions for each operator. 
When twisting the nut with the robot hand (\texttt{hand\_twist}), the forceful kinematic chain composed of the robot joints and the robot's grasp on the nut.
Likewise, when twisting with the spanner (\texttt{tool\_twist}), the chain is composed of the robot joints, the robot's grasp on the spanner and the spanner's grasp on the nut.
Grasps are evaluated via \texttt{(StableGrasp \variable{o}~\variable{g}~\variable{w})} and robot joints are evaluated via \texttt{(StableJoints \variable{a}~\variable{t}~\variable{w})}.
The sampler \texttt{plan-nut-tool} evaluates the stability of the spanner's grasp on the nut.  

We also constrain that the bolt, \variable{o_{b}} is fixtured, using any of the available fixturing methods.
In \apref{appendix:domains} we describe the domain-specific elements for the childproof bottle and vegetable cutting domain.

\subsubsection{Modeling Effort}
We next consider what is required in applying this framework to new settings. 
We consider two categories of modifications: incorporating new assessments of the forceful kinematic chain and fixturing constraints and incorporating new domain-specific elements. 

\sref{sec:kinematicChain} defines several mathematical models for assessing the stability of a joint. 
These models were incorporated as test samplers within PDDLStream. 
Since the framework is agnostic to the Pythonic implementation of the sampler, it is straightforward to swap out various stability models. 

Adding a new joint type, and its corresponding stability model, requires adding both the corresponding sampler and the certified fact.
Adding a new fixturing method may require adding operators to utilize the fixturing method (e.g. incorporating vise fixturing required adding vise actuation actions). 
It also requires identifying and integrating the appropriate stability assessment.  

Critically, once such additions are made, they can be reused across many domains. 
In our experience, this allows domains to build off of each other, decreasing the implementation effort with each new domain.
For example, to model the contact patch between the beam and the table in the nut twisting domain, we incorporated the generalized friction cone. 
Thus, when implementing the cutting domain, we reused that same abstraction to model the contact patch between the vegetable and the table. 


Solving a new task often involves adding domain-specific operators to capture a new action space (e.g. for cutting we had to add the \texttt{slice-cut} operator).
We approach adding a new operator by first defining the preconditions and effects that characterize the kinematics and geometry. 
For example, if an operator uses a tool (as \texttt{tool-twist} does), then a precondition is that the robot must be grasping the tool. 
Next, if the operator is exerting a forceful operation, the relevant forceful kinematic chains must be stable. 
This requires identifying the joints within the chain and the appropriate corresponding stability models and adding the stability checks as preconditions. 

Finally, we must write the sampler that defines the operator's parameterized controller. 
This step involves combining the existing lower-level controllers (joint-space position controller, grasp controller, guarded move controller, cartesian impedance controller, etc.) to create the desired behavior. 
We find that even with operator-specific parameterized controllers, there often is significant overlap across domains with respect to how to combine the low-level controllers.

 
\subsection{Robust Planning}
\label{sec:robust}

Given the ability to generate plans that satisfy motion- and force-based constraints, we now aim to produce \textit{robust} plans.
In particular, we focus on protecting against stability-based failures along the forceful kinematic chains due to \textit{uncertainty} in physical parameters.
For example, we want to discourage the system from selecting a grasp where a small change in the friction coefficient would break the stability of the grasp, leading to the object slipping.

To generate robust plans, we associate each operator with a probability of success $P[success(operator)]$.
As given in \aref{algo:actionProb}, we assess the probability of an operator's success via Monte Carlo estimation, i.e. we draw sets of parameter values, where each parameter is perturbed by some uniformly-sampled epsilon, and evaluate the stability of each forceful kinematic chain.
We perturb, when applicable, parameters such as the friction coefficient, the planned applied wrench, the contact frame and the effective size of the contact patches~\footnote{The details of the perturbations used in the experiments in \sref{sec:robustDemos} are given in \apref{appendix:robust}}.

As an example, \figref{fig:robustLS} visualizes how sampling various grasp parameters essentially generates a large set of possible limit surfaces. By evaluating the stability with respect to all of them, we capture how the grasp stability is impacted by uncertainty.

\begin{algorithm}[t!]
\caption{Compute $P[success(operator)]$}\label{algo:actionProb}
\begin{algorithmic}[1]
\State \textbf{Given:} Chain $c$, wrench $w$, parameters $p$
\State \textbf{Initialize:} $i = 0$
\Procedure{for j=0:n}{}
\State $p_{\epsilon} \gets$ Perturb $p$ by epsilon
\If{\Call{StabilityCheck}{$c, w, p_{\epsilon}$}}
\State $i \mathrel{+=} 1$
\EndIf
\EndProcedure
\State \Return $\frac{i}{n}$
\end{algorithmic}
\end{algorithm}

\begin{figure}[t!]
\centering
    \includegraphics[width=0.6\columnwidth]{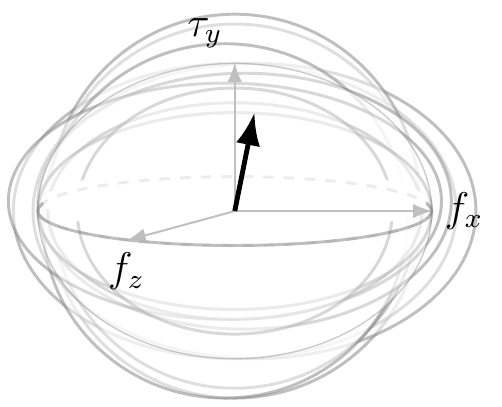}
\caption{By sampling over the friction coefficient, $\mu$, radius of contact $r$ and task wrench, we can assess if a contact is stable in face of uncertainty in those parameters.}  
\label{fig:robustLS}
\end{figure}

We define the cost of an operator as:
\begin{equation}
cost(operator) = -\log(P[success(operator)]).
\label{eq:actionCost}
\end{equation}
The cost of a plan is then the sum over the cost of all the operators in the plan.
Minimizing this cost is equivalent to maximizing the plan success likelihood.

To generate robust plans, we use PDDLStream's \textit{cost-sensitive planning} where, given non-negative, additive operator costs that are functions of the operators's parameters, the planner searches for a plan that is below a user-provided cost threshold.
Cost functions are specified in the domain by adding that the effect of an operator is increasing the total cost of the plan by the value computed by the operator's cost function. 

With this cost definition, the cost threshold corresponds to the probability of succeeding during open-loop execution given uncertainty in the physical parameters.

\section{Empirical Evaluation}
\label{sec:demos}

Using the childproof bottle domain, we demonstrate how the planner finds a wide variety of solutions and how the feasibility of these solutions depends on the environment.
In each of the three domains, we show how accounting for uncertainty by planning robustly leads to the robot making different choices, both with respect to the strategy and with respect to the continuous choices. 

In the supplemental material we include simulation and real robot videos showing a variety of strategies in each of the three domains.  

\subsection{Exploring Strategies}

We demonstrate the range of the plans that the planner generates by considering two settings in the childproof bottle domain (\tref{table:ablation}).
The robot can fixture the bottle using a variety of possible strategies including stably grasping with another robot, using a vise or using frictional contact with the surface, where the surface is either the table or a high-friction rubber mat~\footnote{In this domain we do not consider the fixturing method of weighing down the bottle, since this is not geometrically feasible.}. 
The robot can apply the push-twist operation through a variety of contacts: a grasp, fingertips, a palm or a grasped pusher tool. 

In the first setting, we search over several possible fixturing strategies, fixing the push-twisting strategy to use a grasp contact.
In this setting the bottle starts at a random location on the table. 
In the second setting, we search over all possible push-twisting strategies. 
In this setting the bottle starts on a high-friction rubber mat in order to leverage the friction of the surface for fixturing. 

Because the underlying search over strategies in PDDLStream biases towards plans with the fewest actions, we incrementally invalidate the shorter strategies once found by the planner in order to force exploration of the alternative, longer strategies. 
For example, in the fixturing setting we invalidate fixturing via a robot grasp as a feasible strategy by removing the second arm from the environment. 

\begin{table}[t!]
 \centering
 \begin{tabular}{l c c}
 \hline 
 \multicolumn{3}{c}{Fixturing Ablation} \\ [0.5ex] 
 \hline\hline
 Fixturing Method & \# Steps & Planning Time  (SE) \\
 \hline
 Surface(Table) * & 4 & 177 (51) \\ 
 \hline
 Robot Grasp & 6 & 60 (9.4) \\ 
 \hline
 Surface(Mat)** & 8 & 142 (73) \\ 
 \cline{1-3}
 Vise grasp** & 9 & 95 (35) \\ 
 \hline\hline 
 \multicolumn{3}{c}{Push-Twisting Ablation} \\ [0.5ex] 
 \hline
 Pushtwist Method & \# Steps & Planning Time (SE) \\ 
 \hline\hline
 Grasp & 4 & 37 (1.8) \\ 
 \hline
 Palm & 4 & 25 (3.1) \\ 
 \hline 
 Fingertip* & 4 & 63 (35) \\ 
 \hline
 Pusher Tool** & 8 & 40 (5.1) \\ 
 \hline
\end{tabular}
\caption{For each setting, we provide the number of steps for each strategy and the average planning time in seconds (and standard error) over five runs. *: Utilized a higher friction coefficient $\mu$ to increase feasibility **: Invalidated shorter strategies to force to planner to find these longer strategies.}
\label{table:ablation}
\end{table}

\subsubsection{Fixturing Setting}
We first consider searching over various fixturing strategies. 
Looking first for the strategy with the fewest number of actions, the planner first tries to fixture the bottle against the table surface by applying additional downward force. 
However, the friction coefficient between the table and bottle is small enough that this is not a viable strategy: even when applying maximum downward force the robot cannot fixture the bottle. 
So instead, the planner fixtures using the second robot to grasp the bottle. 

Removing the second arm from the environment forces the planner to discover new strategies. 
One strategy employed is to use a pick-and-place operation to move the bottle to a high-friction rubber mat, where it is possible to sample enough additional downward force such that the bottle is fixtured. 
Another strategy is to use a pick-and-place operation to move the bottle into the vise, where it can be fixtured. 

\subsubsection{Push-Twisting Setting}
We next consider searching over various push-twisting strategies. 
Three of the twisting strategies, contacting via a grasp, the fingertips or the palm, are of equivalent length, and are therefore equally attempted when searching over strategies. 
We can view the viability of finding successful parameters to these actions, and thus probability of employing that strategy, as how easy it to sample satisfying values.

For example, twisting the cap by pushing down with the palm is only stable if, given the values of the radius of the palm and the friction coefficient, 
the system can exert enough additional downward force. 
Decreasing either the radius or the friction coefficient narrows the space of feasible downward forces. 
Since the fingertips have a smaller radius, as compared to the palm, it is harder to sample a set of satisfying values for the fingertips. \\ 
\newline
In each of these settings, we demonstrate that the planner finds a variety of different strategies and that the choice of strategy adapts to what is feasible in the environment. 
This adaptability allows the planner to generalize over a wide range of environments. 

\subsection{Generating Robust Plans}
\label{sec:robustDemos}


We next examine how leveraging cost-sensitive planning enables the robot to generate more robust plans. 

\subsubsection{Childproof Bottle Domain}

\begin{figure}[t!]
\centering
    \includegraphics[width=0.48\columnwidth]{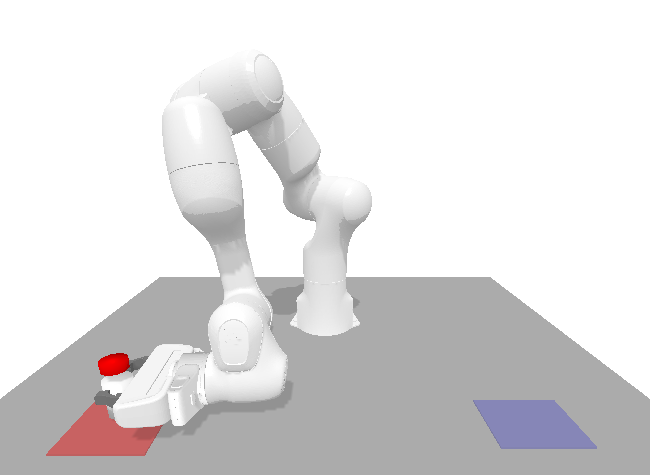}
    \includegraphics[width=0.45\columnwidth]{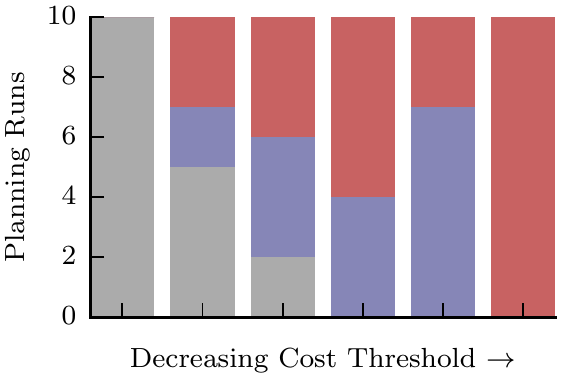}
\caption{In opening the childproof bottle, the robot can fixture against the low-friction table (\protect\markerGray), a medium-friction mat (\protect\markerBlue) or a high-friction mat (\protect\markerRed). For each cost threshold we run the planner ten times, noting which surface is used. As the cost threshold decreases, the robot is forced to more frequently use higher friction surfaces that are more robust to uncertainty.}
\label{fig:robustMat}
\end{figure}

\begin{table}[t!]
 \centering
 \begin{tabular}{c c}
 \hline
 Cost Threshold & Planning Time  (SE) \\
 \hline\hline
 $\infty$ & 6.9 (0.1) \\ 
 \hline
 0.5 & 41 (14) \\
 \hline
 0.4 & 69 (17) \\ 
 \hline
 0.3 & 77 (10) \\ 
 \hline
 0.2 & 94 (18) \\
 \hline 
 0.1 & 127 (29) \\ 
 \hline
\end{tabular}
\caption{We evaluate how decreasing the cost threshold impacts what fixturing surface the planner uses in the childproof bottled domain (\figref{fig:robustMat}). As the cost threshold decreases, the planning time increases.}
\label{table:timeRobustness}
\end{table}
\arrayrulecolor{black}

\begin{figure*}[t!]
\centering
    \includegraphics[width=0.36\textwidth]{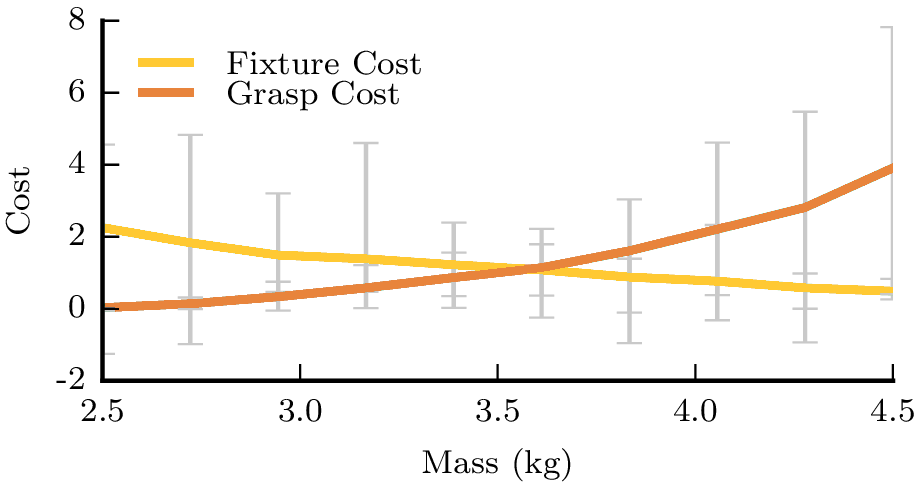} \hspace{1cm}
    \includegraphics[width=0.25\textwidth]{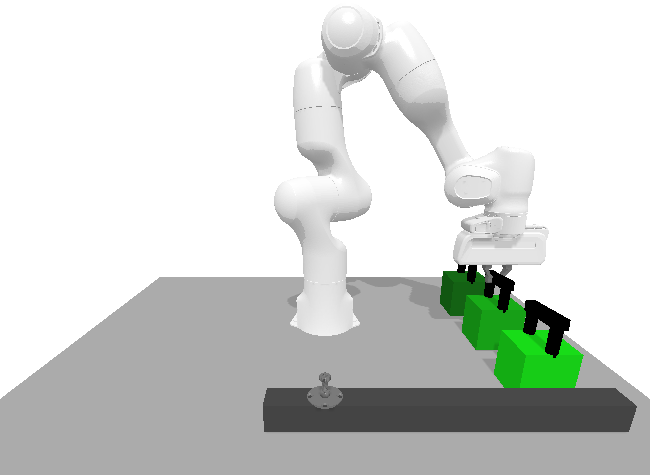} \hspace{1cm}
    \includegraphics[width=0.25\textwidth]{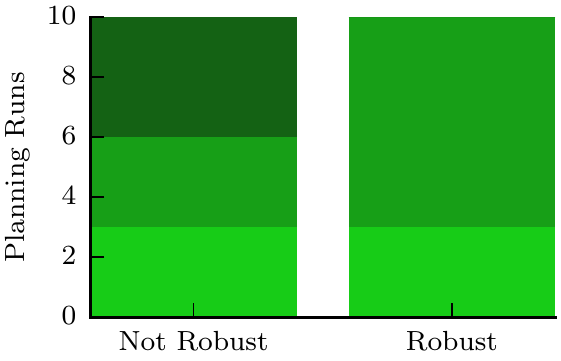}
\caption{In the nut-twisting domain we consider the trade-off between the grasp cost and the fixturing cost. On the left, at each weight value, we randomly sample, 100 times, the pose of the weight along the beam and the grasp on the weight. Since, at the extremes, some costs evaluate to infinity, we plot the median and a 95\% confidence interval. We then demonstrate how the trade off impacts the choices made by the planner by considering an environment in which there are three possible masses, as shown in the center. The robot can fixture using the 2.6kg mass (\protect\markerLow), the 3.5kg mass (\protect\markerMedium) or the 4.4kg mass (\protect\markerHigh). Without accounting for robustness, the robot chooses any of the masses. When planning robustly, the robot more often picks the medium weight, which balances the trade-off in costs. In both cases we run the planner ten times, noting which weight is used.}
\label{fig:robustWeight}
\end{figure*}

In the childproof bottle domain, we explore how accounting for robustness impacts what surface the robot uses to fixture against. 
As visualized in \figref{fig:robustMat}(left), the robot can choose between fixturing on three surfaces: the low-friction table, the medium-friction blue mat or the high-friction red mat. 
For all planning instances we start the bottle on the table and set the friction coefficient between the table and the bottle to be just high enough to make fixturing feasible.
At each cost threshold, we run the planner ten times.  

If the planner does not account for robustness, the robot fixtures with the table every time.
This is because doing so is feasible and results in the shortest plan. 

When considering robustness, the planner evaluates that fixturing using the table produces a feasible but brittle plan, resulting in a high cost plan that is unlikely to succeed.
To avoid this, the planner completes a pick-and-place action to relocate the bottle to one of the two mats that have a higher friction coefficient and thus offers a more robust fixturing surface. 
As we decrease the cost threshold, the planner is forced to use exclusively the high-friction red mat.

\tref{table:timeRobustness} shows how decreasing the cost threshold increases the planning time.
The choice of the cost threshold can therefore be viewed as trading off between the probability of the plan's success and the time to find the plan.

\subsubsection{Nut Twisting Domain}

In the nut-twisting domain we explore robustness by considering a scenario where the robot must choose between several weights, of varying mass, to fixture the beam with.

First, for a given mass, we sample 100 placement locations along the beam holding the bolt and evaluate two robustness metrics: how robustly the weight fixtures the beam and how robustly the robot is able to grasp (and therefore move) the weight to this placement. 
\figref{fig:robustWeight}-left shows the trade-off: a heavier weight more easily fixtures the beam but is harder to grasp robustly. 
In finding a robust plan, and hence a low-cost plan, the planner is incentivized to act like Goldilocks and pick the weight that best balances this trade-off. 

We can see this in action when running the planner in a setting with three weights of various masses (2.6, 3.5, 4.4 kg), shown in \figref{fig:robustWeight}-center, where a darker color corresponds to a larger mass.
\figref{fig:robustWeight}-right shows that when the planner does not account for robustness, the three weights are selected equally, since all can be used to produce feasible plans. 
However, when accounting for robustness the planner more often selects the medium weight, which balances the trade off between the cost functions. 
In both instances, the planner was run ten times. 

\subsubsection{Vegetable Cutting Domain}

\begin{figure}[t!]
\centering
 \begin{tikzpicture}[box/.style={rectangle, align=center}]
   \node (chart) {\includegraphics[width=0.9\columnwidth]{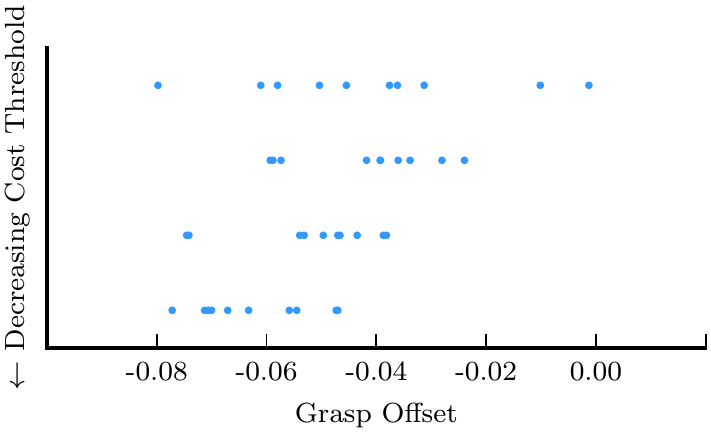}};
   \node (knife) [above=of chart, yshift=-1.3cm, xshift=-0.3cm] {\includegraphics[width=0.8\columnwidth]{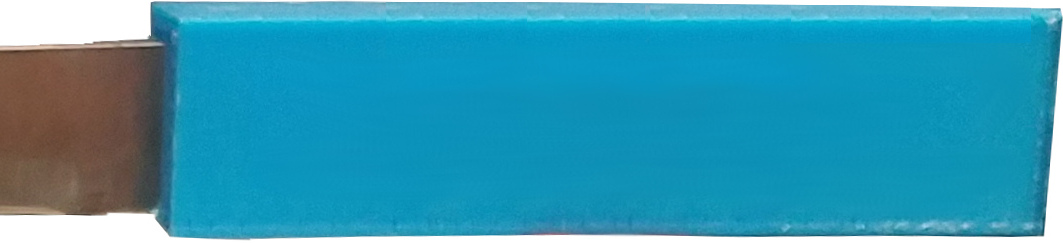}}; 
  \end{tikzpicture}
\caption{In the vegetable cutting domain, we show how robust planning leads the planner to select grasps that are closer to the blade of the knife, because doing so creates a smaller torque that the grasp needs to resist. For each cost threshold, the planner is run 10 times and the grasping offset is plotted. An offset of $0$ corresponds to a grasp at the butt of the knife.}
\label{fig:robustKnife}
\end{figure}

In the vegetable cutting domain, we explore how planning robustly impacts the continuous choice of the grasp on the knife. 

We return to the two possible grasps on the knife shown in \figref{fig:graspCompareExamples}.
The side grasp, shown on the bottom, relies on normal reaction forces to resist the torque experienced while exerting the downward force, the first half of the cutting action. 
As such, the grasp is very stable and robust, regardless of the location of the grasp along the length of the handle (the annotated $x$-axis). 
In contrast, the top grasp (shown in \figref{fig:graspCompareExamples}(top)) relies on frictional forces to resist the exerted torque. 
The stability of the top grasp varies based on its location along the handle and by evaluating the robustness of these grasps, we can empower the planner to make a more informed choice. 

To demonstrate this, we restrict our grasp set to only those that grasp the knife from the top (i.e. like \figref{fig:graspCompareExamples}(top)). 
Additionally, we consider cutting a softer object, reducing the magnitude of the downward force ($f_{z, 0}$) such that it is feasible to find a stable grasp.
Specifically, by default we set $f_{z, 0}=5N$, while for the soft object we use $f_{z, 0}=3N$.We refer to this as the high force setting and low force setting, respectively.

Across several cost thresholds, considering only the cost with respect to the grasp stability during the downward cut, we plot the location of the grasp along the handle in \figref{fig:robustKnife}.
For each cost threshold, the planner was run ten times.
When the planner does not account for robustness, shown at the top of \figref{fig:robustKnife}, the robot selects any grasp along the handle. 
Accounting for robustness, as the cost threshold decreases, the planner selects grasps that are closer to the blade and thus create a smaller lever arm, decreasing the amount of torque the grasp must be stable with respect to.

We next evaluate how this choice impacts the robot's ability to successfully slice, by testing on three different foods of varying hardness: cucumbers, bananas with the peel and bananas without the peel.  
From the continuous grasp set, we use three representative grasps, shown in \tref{table:robustCutting}: grasping the side of the handle (\texttt{side\_grasp}), grasping from the top of the handle close to the blade (\texttt{top\_grasp\_close}) and grasping from the top of the handle cloes to the butt of the knife (\texttt{top\_grasp\_far}). 

For each grasp and each food we execute the same downward force-motion, without evaluating stability or robustness. 
For the cucumber and the banana with the peel we repeat this fifteen times. 
For the banana without the peel we repeat this ten times. 
\tref{table:robustCutting} classifies the results into three categories: success (fully slicing through the object), partial success (slicing through at least half of the object) and failure (not significantly slicing the object).

While we cannot precisely know the required downward force ($f_{z, 0}$) for each food, we qualitatively assess the results. 
From our experience, the banana with the peel requires a force slightly higher than the low force setting, as considered in \figref{fig:robustKnife}. 
As stated previously, a planner that does not account for robustness would consider \texttt{top\_grasp\_close} and \texttt{top\_grasp\_far} equally while a robust planner would prefer \texttt{top\_grasp\_close}.
The \texttt{top\_grasp\_far} cannot slice through the banana with the peel (all fifteen attempts are failures) while the \texttt{top\_grasp\_close} is able to cut through, at least partly, the majority of the time. 
In contrast, the banana without the peel corresponds to a soft enough object such that all grasps can successfully slice through.

Finally, we estimate that the cucumber requires the high force setting. 
As predicted in \figref{fig:graspCompareExamples}, the \texttt{side\_grasp} is sufficiently stable to slice through the object.
However neither \texttt{top\_grasp\_close} nor \texttt{top\_grasp\_far} succeed because for both, the forceful kinematic chain breaks (the knife moves considerably within the robot's end effector). 
While the robust planner, considering only grasps from the top, evaluates the \texttt{top\_grasp\_close} to be robust enough with respect to the low force setting, this is is inadequate for the cucumber, whose true required force is significantly larger. 

As a final note, one might wonder why we do not fix the planner to always use \texttt{side\_grasp}, since \tref{table:robustCutting} demonstrates it to be the most robust grasp. 

First, we currently use a generic uninformed grasp sampler that is common across all of the handled tools (the pusher tool in the childproof bottle domain, spanner in the nut twisting domain and knife in cutting domain).
Prescribing a preferred grasp for each tool and task, via an informed grasp sampler, would both increase the modeling effort and decrease the generalization.   
Instead, with an uninformed sampler, we allow the planner to reason over the best grasp for the specific scenario. 
As shown in \tref{table:timeRobustness}, this is at some computational cost. 

Second, while the side grasp may seem to be the best grasp with respect to this constraint, in the context of a multi-step manipulation problem, there are many, often competing, constraints. 
While for simplicity we have focused on evaluating the grasp with respect to the downward force, the grasp must also be stable with respect to the translational slice. 
Additionally, as shown in \figref{fig:cutgrid}, \figref{fig:removeBlock} and \figref{fig:graspCompareExamples}, due to the geometry of the robot's end effector, \texttt{side\_grasp} is only collision-free if the object is significantly raised. 
For foods where such a grasp is not strictly necessary, the planner can select a more reachable grasp. 
This flexibility is critical to enabling the planner to adapt to a variety of different environments.

\begin{table}[t!]
 \centering
\begin{tabular}{ c c | c | c}
   & \multicolumn{1}{c}{\includegraphics[width=0.15\columnwidth]{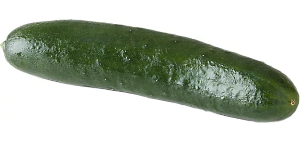}} & 
     \multicolumn{1}{c}{\includegraphics[width=0.15\columnwidth]{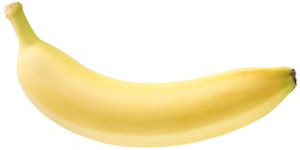}} & 
     \multicolumn{1}{c}{\includegraphics[width=0.15\columnwidth]{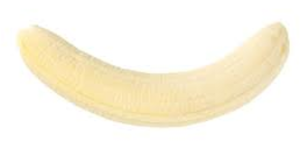}} \\
 \begin{minipage}{0.24\columnwidth}\includegraphics[width=\columnwidth]{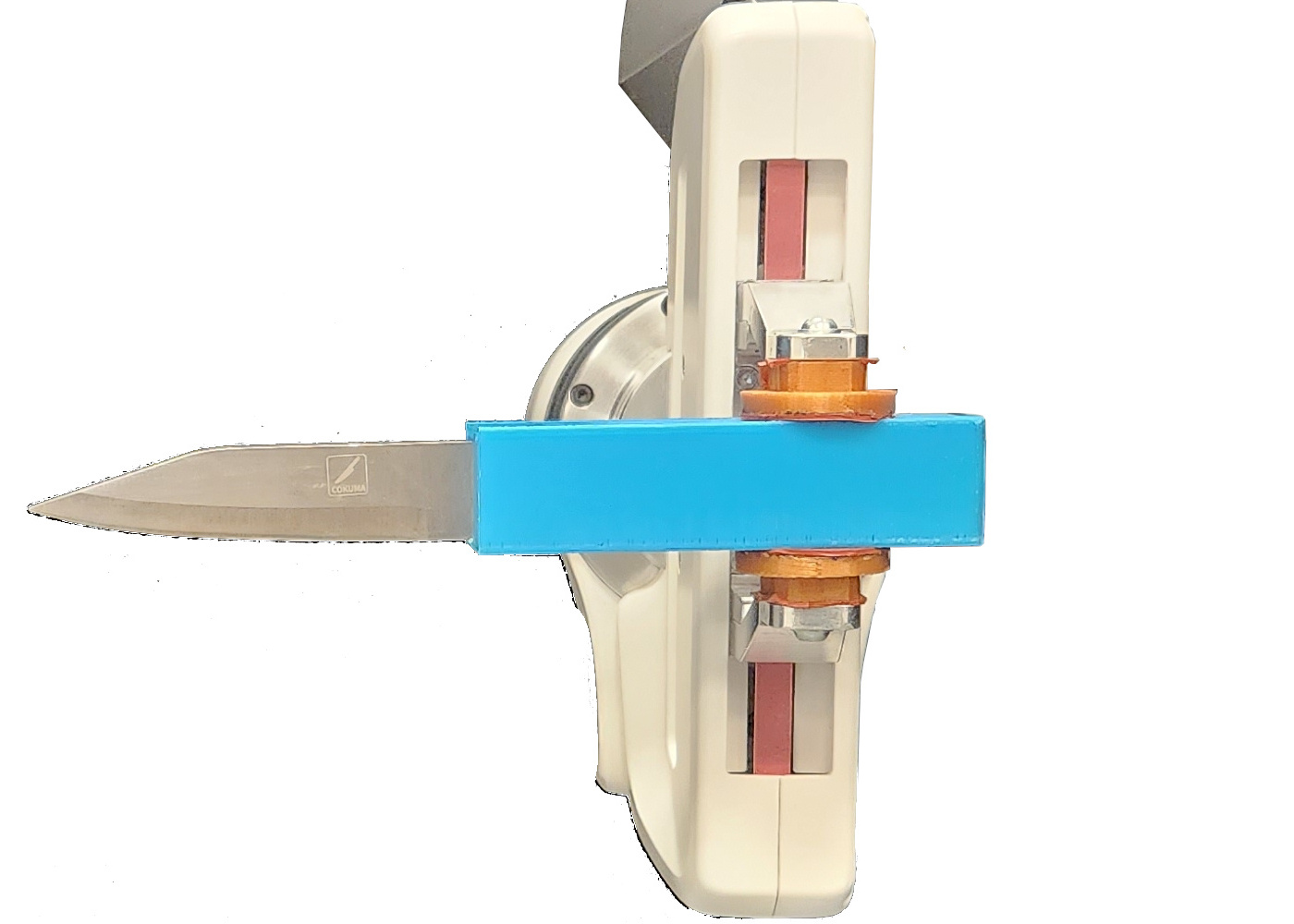}\end{minipage} & 6/9/0 & 7/6/2 & 10/0/0 \\
 \cline{2-4}
 \begin{minipage}{0.24\columnwidth}\includegraphics[width=\columnwidth]{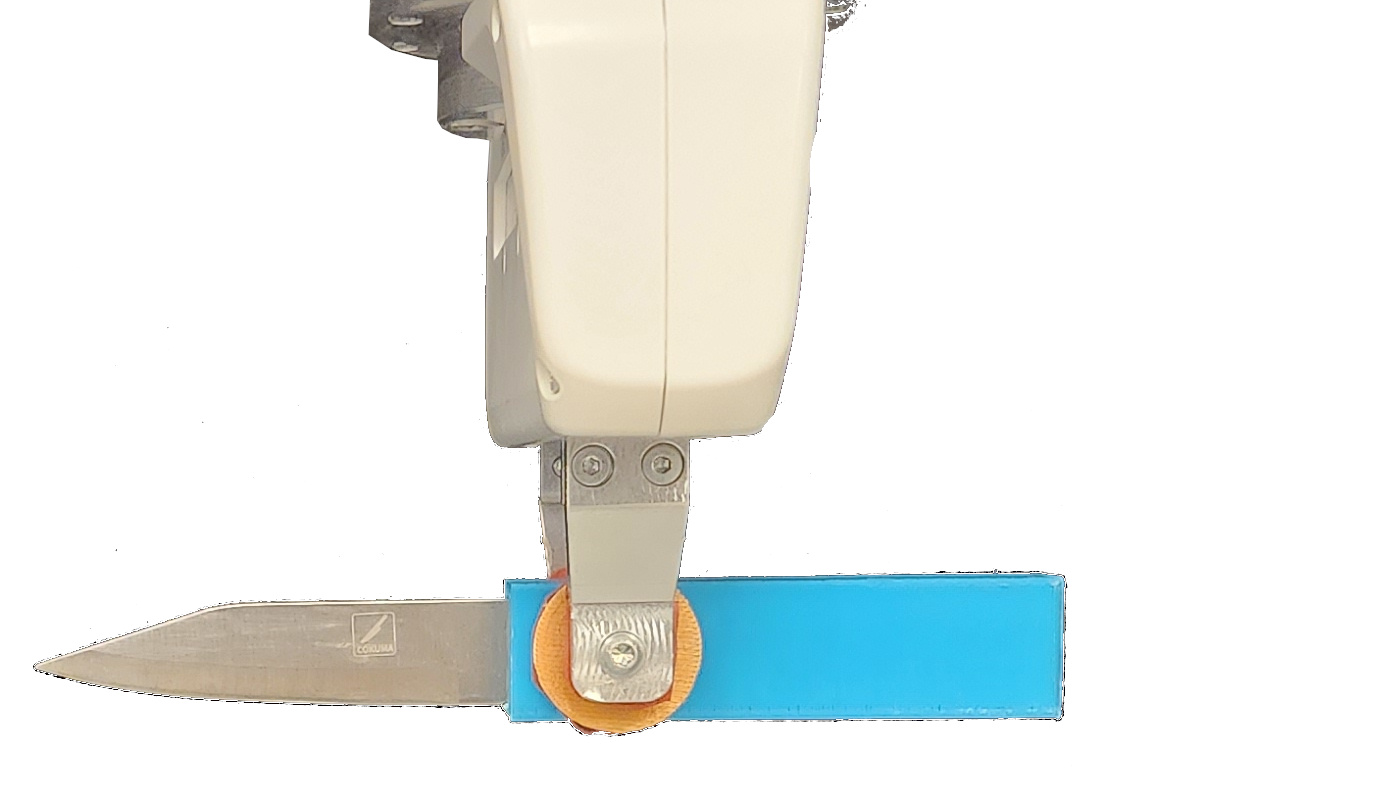}\end{minipage} & 0/0/15 & 3/8/4 & 10/0/0 \\
 \cline{2-4}
 \begin{minipage}{0.24\columnwidth}\includegraphics[width=\columnwidth]{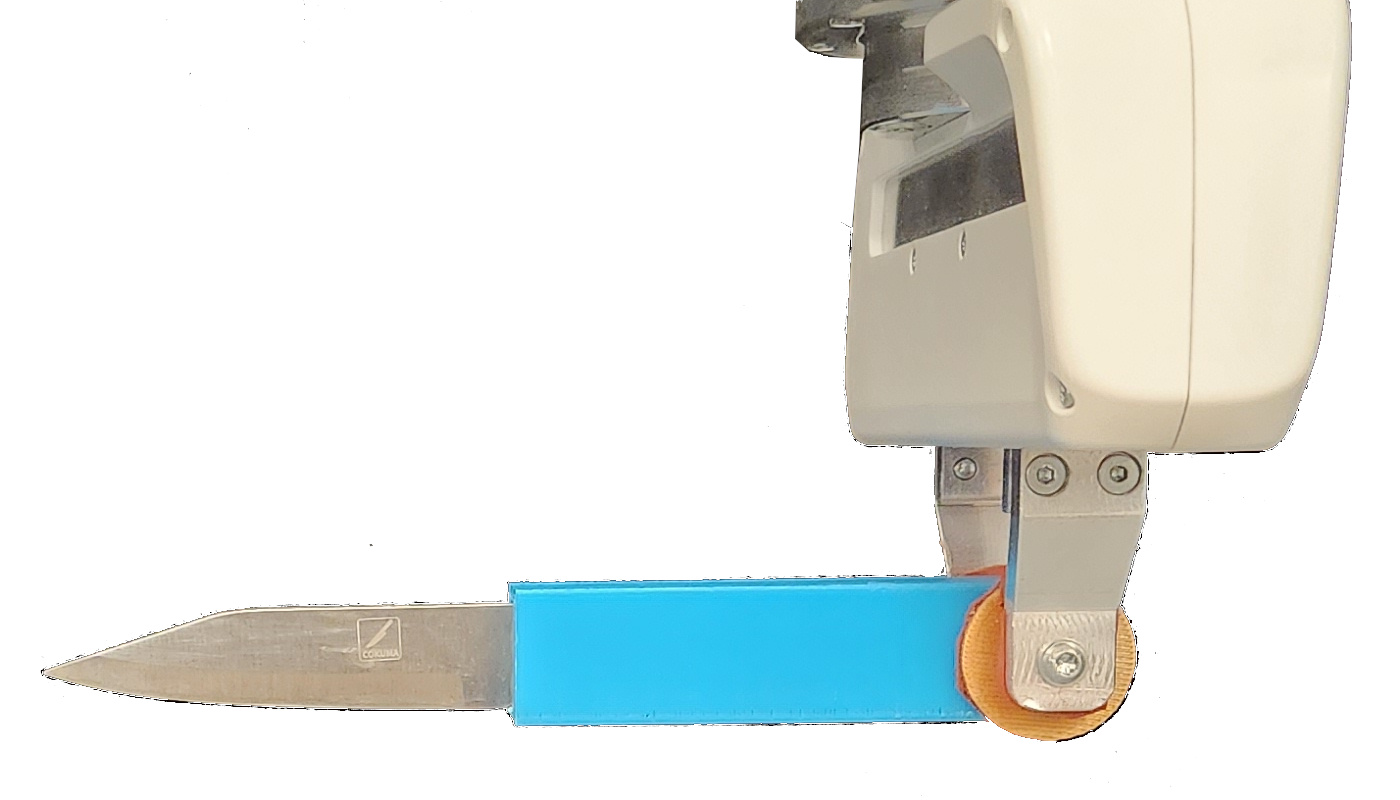}\end{minipage} & 0/0/15 & 0/0/15 & 9/1/0 \\
\end{tabular}
	\caption{We evaluating slicing success for three grasps (from top to bottom: \texttt{side\_grasp}, \texttt{top\_grasp\_close} and \texttt{top\_grasp\_far}) across three foods. For the cucumbers and bananas with peel we perform 15 iterations, for the banana without the peel we perform 10. We classify each interaction as a success / partial success / failure.}
\label{table:robustCutting}
\end{table}
\arrayrulecolor{black}

\section{Discussion}
\label{sec:discussion}

This paper proposes a planning framework for solving forceful manipulation tasks. 
We define forceful manipulation as a class of multi-step manipulation tasks that involve reasoning over and executing forceful operations, where forceful operations are defined as the robot applying a wrench at a pose.

\subsection{Summary of Contributions}

Solving forceful manipulation tasks requires planning over a hybrid space of discrete and continuous choices that are coupled by force and motion constraints. 
We frame the primary force-related constraint as the system's ability to stably exert the desired task wrench.
To capture this, we propose the forceful kinematic chain constraint which evaluates if every joint in the chain is stable under the application of the imparted wrench and gravity. 
For each class of joint in the chain, which may be robot joints, grasps or frictional contacts, we discuss a model for evaluating its stability. 

In addition to the forceful kinematic chain constraint, the planner must reason over other force-related constraints, such as the requirement to fixture objects, and over motion constraints, such as the requirement to find collision-free paths. 
To plan multi-step sequences that respect these constraints, we augment an existing task and motion planning framework, PDDLStream~\citep{garrett2020pddlstream}. 
We illustrate our system in three example domains: opening a push-and-twist childproof bottle, twisting a nut on a bolt and cutting a vegetable.

While PDDLStream finds a satisficing plan, the plan may not be robust to uncertainty.
To find robust plans, we propose using cost-sensitive planning to select actions that are robust to perturbations. 
We specifically focus on uncertainty in the physical parameters that determine the stability of the forceful kinematics chains.
Our demonstrations show how cost-sensitive planning enables the robot to make more robust choices, both with respect to the strategy, such as what fixturing method to use, and the continuous choices, such as which grasp to pick. 


\subsection{Future Directions}

In this work we assume that the planner has access to many different parameters, such as the magnitude and form of the forceful operation, object models and object poses. 
We also assume the domain is given in the form of the fact types and lifted operators, with their preconditions, effects and controllers. 
Techniques from machine learning, combined with this planning framework, could be used to relax these assumptions and enable wider generalizations~\citep{konidaris2018skills,wang2021learning, silver2021learning, liang2022search}.

\bibliographystyle{plainnat}
\bibliography{references}

\appendix
\section{Implementation Details}
\label{appendix:implementation}

We provide some implementation details with respect to the controller used to exert wrenches and the specific motion planning and grasping tools used in the kinematic samplers. 



\subsection{Controller for Exerting Wrenches}
\label{appendix:controller}

Each of the operators that apply the forceful operation, e.g. \texttt{hand\_twist} and \texttt{tool\_twist} in the nut twisting domain, are associated with a controller that must exert a wrench. 
While there are several control methods for exerting wrenches explicitly, such as force control~\citep{zeng1997overview} or hybrid position-force control~\citep{mason1981compliance,de1988compliant,hou2019robust}, we opt to use a Cartesian impedance controller.
Cartesian impedance control regulates the relationship between force and motion by treating the interplay of interaction forces and motion derivatives as a mass-spring-damper system~\citep{hogan1985impedance, albu2003cartesian,ott2008cartesian}. 

\begin{figure}[t!]
\centering
   \includegraphics[width=\columnwidth]{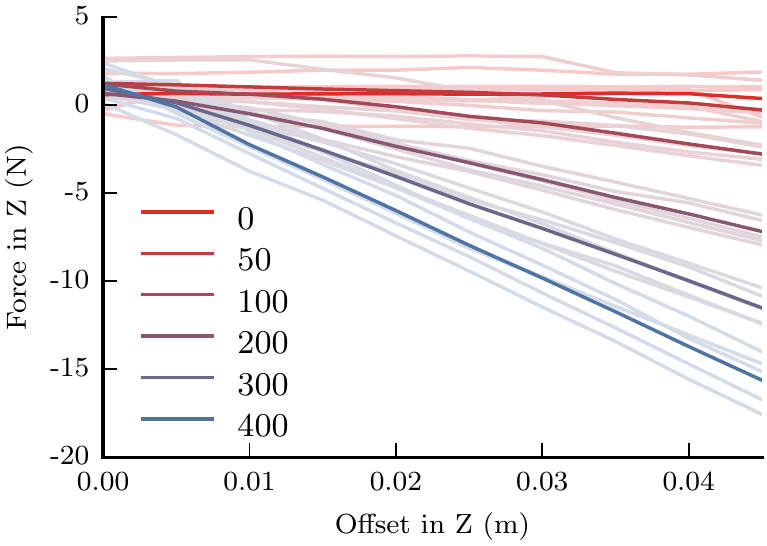}
   \caption{Across varying stiffnesses with Cartesian impedance control, we plot the experimental relation between the offset in the commanded offset in the z direction and the exerted force in z, as measured by an external force-torque sensor. For each stiffness we plot all five experimental runs, bolding the average. The result shows that the relation between the offset and force exerted is nearly linear.}
\label{fig:impedanceExperiments}
\end{figure}

We can exert a wrench by offsetting the target Cartesian pose to be below the point of contact and adjusting the impedance parameters~\citep{kresse2012movement}.
Intuitively this exerts a wrench by compressing the spring rendered by the robot. 
We chose to only vary the stiffness matrix, $K_{p}$, and set the damping matrix, $K_{d}$, to be approximately critically damped, i.e. $K_{d} = 2\sqrt{K_{p}}$.
As shown in \figref{fig:impedanceExperiments} we can experimental characterize the relationship between the exerted wrench, the pose offset and the stiffness.
The planner uses this experimental relation to select, given the stiffest possible setting, the desired pose offset. 
Thus, each of the samplers that generate Cartesian impedance paths must generate the series of setpoints in $SE(3)$, accounting for the pose offset, and the stiffnesses.

\subsection{Motion Planning and Grasping}

Collision-free motion planning, such as what is used in samplers \texttt{plan-free-motion} and \texttt{plan-holding-motion}, can be implemented with any choice of motion planner. 
We use a python implementation of bidirectional RRT (BiRRT)~\citep{lavalle2001rapidly}. 
In these settings, a considerable fraction of the overall PDDLStream planning time is spent performing motion planning. 
A more efficient implementation or algorithm would significantly speed up the reported results. 

Any grasp generation method could be used to generate grasps for samplers such as \texttt{sample-grasp}. 
For each graspable object we define the grasp set using Task Space Regions (TSRs), which are a compact method for describing pose constraints that allows for random sampling~\citep{berenson2011task}. 

\subsection{Robust Planning Experimental Parameters}
\label{appendix:robust}

As detailed in \sref{sec:robust}, to generate robust plans we assess the probability of a forceful kinematic chain being stable via Monte Carlo estimation.
Here we briefly describe how each parameter was perturbed for the experiments in \sref{sec:robustDemos}. 

The friction coefficient is additively perturbed by a value uniformly sampled between $[-0.1, 0.1]$.
The applied wrench is multiplicatively perturbed by a value uniformly sampled between $[0.5, 1.5]$.
The contact frame is additively perturbed by a value uniformly sampled between $[-5mm, 5mm]$ if the frictional joint is formed by a robot grasp and $[-10mm, 10mm]$ otherwise. 
The contact patch size ($r$, in the context of the ellipsoidal limit surface) is recomputed. 

\section{Further Domain Specification}
\label{appendix:domains}

In \sref{sec:forceplanning} we detailed the lifted operators, fact types and samplers used for the nut-twisting domain. 
The domain was composed of elements from the pick-and-place domain, the forceful manipulation additions and domain-specific additions. 
In this section we first detail more forceful manipulation additions before providing the domain-specific elements for the childproof bottle and vegetable cutting domains.

\subsection{Forceful Manipulation Additions}
\label{appendix:fm_adds}


\begin{table*}[t!]
\begin{subtable}{\textwidth}
 \centering
 \begin{tabular}{l c c}
   \hline
   Action & Preconditions & Effects \\ [0.5ex]
   \hline\hline
   open\_vise & 
    \makecell[cc]{(ViseHand $v$) ($\neg$ (Movable $o$)) $\wedge$ (AtPose $o~p_{o}$) $\wedge$ \\
                  (AtGrasp $v~o~g_{o}$) $\wedge$ \textcolor{samplerGreen}{(ViseMotion $v~o~p_{o}~g_{o}~w~t$)}} & 
    \makecell[cc]{(Movable $o$) \\ ($\neg$ (AtGrasp $v~o~g_{o}$))} \\
   \hline 
   close\_vise & 
    \makecell[cc]{(ViseHand $v$) $\wedge$ (Movable $o$) $\wedge$ (AtPose $o~p_{o}$) $\wedge$ \\ (On $o~v$) $\wedge$ 
                  \textcolor{samplerGreen}{(ViseMotion $v~o~p_{o}~g_{o}~w~t$)}} & 
    \makecell[cc]{(AtGrasp $v~o~g_{o}$) \\ ($\neg$ (Movable $o$))} \\
   \hline
   pushin\_tool & 
    \makecell[cc]{(AtConf $a~q_{0}$) $\wedge$ (AtPose $o~p_{o}$) $\wedge$ (Pusher $o_{p}$) $\wedge$ \\
                  (AtGrasp $a~o_{p}~g_{p}$) $\wedge$ 
                  \textcolor{samplerGreen}{(StableGrasp $o_{p}~g_{p}~w$)} $\wedge$ \\
                  \textcolor{samplerGreen}{(SampleWrench $w~o$)} $\wedge$ \\
                  \textcolor{samplerGreen}{(ContactToolMotion $a~o_{p}~g_{p}~o~p_{o}~p_{oc}~w~q_{0}~q_{1}~t$)} $\wedge$ \\
                  ($\neg$ (TrajUnsafe $a~t$))} & 
    \makecell[cc]{(InContact $o~o_{p}~c~p_{oc}~w$) \\ ($\neg$ (AtConf $a~q_{0}$)) \\ (AtConf $a~q_{1}$)} \\
   \hline 
   pushout\_tool & 
    \makecell[cc]{(AtConf $a~q_{0}$) $\wedge$ (AtPose $o~p_{o}$) $\wedge$ (Pusher $o_{p}$) $\wedge$ \\ 
                  (AtGrasp $a~o_{p}~g_{p}$) $\wedge$ 
                  (InContact $o~o_{p}~c~p_{oc}~w$) $\wedge$ \\
                  \textcolor{samplerGreen}{(ContactToolMotion $a~o_{p}~g_{p}~o~p_{o}~p_{oc}~w~q_{0}~q_{1}~t$)} $\wedge$ \\
                  ($\neg$ (TrajUnsafe $a~t$))} & 
    \makecell[cc]{($\neg$ (InContact $o~o_{p}~c~p_{oc}~w$)) \\ ($\neg$ (AtConf $a~q_{0}$)) \\ (AtConf $a~q_{1}$)} \\
   \hline 
   pushin\_rcontact & 
    \makecell[cc]{(AtConf $a~q_{0}$) $\wedge$ (AtPose $o~p_{o}$) $\wedge$ \\
                   (RContact $c$) $\wedge$ 
                  \textcolor{samplerGreen}{(SampleWrench $w~o$)}  $\wedge$ \\
                  \textcolor{samplerGreen}{(ContactMotion $a~c~o~p_{o}~p_{oc}~w~q_{0}~q_{1}~t$)} $\wedge$ \\ 
                  ($\neg$ (TrajUnsafe $a~t$))} & 
    \makecell[cc]{(InContact $o~o_{p}~c~p_{oc}~w$) \\ ($\neg$ (AtConf $a~q_{0}$)) \\ (AtConf $a~q_{1}$)} \\
   \hline
   pushout\_rcontact & 
    \makecell[cc]{(AtConf $a~q_{0}$) $\wedge$ (AtPose $o~p_{o}$) $\wedge$ \\
                   (RContact $c$) $\wedge$
                  (InContact $o~o_{p}~c~p_{oc}~w$) \\
                  \textcolor{samplerGreen}{(ContactMotion $a~c~o~p_{o}~p_{oc}~w~q_{0}~q_{1}~t$)}  $\wedge$ \\
                  ($\neg$ (TrajUnsafe $a~t$)) $\wedge$ }  & 
    \makecell[cc]{($\neg$ (InContact $o~o_{p}~c~p_{oc}~w$)) \\ ($\neg$ (AtConf $a~q_{0}$)) \\ (AtConf $a~q_{1}$)} \\
 \end{tabular}
\caption{Actions} \label{table:fm_Actions}
\end{subtable}
\vspace{0.5cm}
\begin{subtable}{\textwidth}
 \centering
 \begin{tabular}{l l}
   \hline
   Derived Fact & Definition \\ 
   \hline\hline
   (ViseFixtured $o~w$) & 
    \makecell[cc]{$\exists$ $v~g_{o}$ (ViseHand $v$) $\wedge$ (AtGrasp $v~o~g_{o}$) $\wedge$ 
                                      \textcolor{samplerGreen}{(StableViseGrasp $o~g_{o}~w$)}} \\ 
   \hline
   (SurfaceFixtured $o~w$) &
     \makecell[cc]{$\exists$ $p_{o}~c~p_{oc}~r$ (AtPose $o~p_{o}$) $\wedge$ (On $o~r$) $\wedge$ 
                                                (InContact $o~c~p_{oc}~w_{e}$) $\wedge$ \\
                                                \textcolor{samplerGreen}{(StableObjContact $c~p_{oc}~o~w_{e}~w$)} $\wedge$
                                                \textcolor{samplerGreen}{(StableSurfaceContact $o~p_{o}~r~w_{e}~w$)}} \\
 \end{tabular}
\caption{Derived Facts} \label{table:fm_Derived}
\end{subtable}
\vspace{0.5cm}
\begin{subtable}{\textwidth}
 \centering
 \begin{tabular}{l c c l}
   \hline
   Sampler & Inputs & Outputs & Certified Facts \\
   \hline\hline
   sample-push-wrench & $o$ & $w$ & 
       \makecell[cc]{\textcolor{samplerGreen}{(SampleWrench $w~o$) $\wedge$ (Wrench $w$)}} \\ 
   \hline
   plan-vise-motion & $v~o~p_{o}~g_{o}~w$ & $t$ &
       \makecell[cc]{\textcolor{samplerGreen}{(ViseMotion $v~o~p_{o}~g_{o}~w~t$) $\wedge$ (Traj $t$)}} \\
   \hline
   plan-tool-contact & $a~o_{p}~g_{p}~o~p_{o}~p_{oc}~w$ & $q_{0}~q_{1}~t$ &
       \makecell[cc]{\textcolor{samplerGreen}{(ContactToolMotion $a~o_{p}~g_{p}~o~p_{o}~p_{oc}~w~q_{0}~q_{1}~t$) $\wedge$} \\ 
                     \textcolor{samplerGreen}{(Conf $q_{0}$) (Conf $q_{1}$) (Traj $t$)}} \\
   \hline
   plan-rcontact-contact & $a~c~o~p_{o}~p_{oc}~w$ & $q_{0}~q_{1}~t$ & 
       \makecell[cc]{\textcolor{samplerGreen}{(ContactMotion $a~c~o~p_{o}~p_{oc}~w~q_{0}~q_{1}~t$) $\wedge$} \\
                     \textcolor{samplerGreen}{(Conf $q_{0}$) $\wedge$ (Conf $q_{1}$) $\wedge$ (Traj $t$)}} \\
   \hline
   test-vise-grasp-stable & $v~o~g_{o}~w$ & & 
       \makecell[cc]{\textcolor{samplerGreen}{(StableViseGrasp $o~g~w$)}} \\
   \hline
   test-contact-stable & $c~p_{oc}~o~w_{e}~w$ & & 
       \makecell[cc]{\textcolor{samplerGreen}{(StableObjContact $c~p_{oc}~o~w_{e}~w$)}} \\
   \hline
   test-surface-stable & $o~~p_{o}~r~w_{e}~w$ & & 
       \makecell[cc]{\textcolor{samplerGreen}{(StableSurfaceContact $o~p_{o}~r~w_{e}~w$)}} \\ 
 \end{tabular}
\caption{Samplers} \label{table:fmSamplers}
\end{subtable}
\caption{The lifted operators, derived facts and samplers common across forceful manipulation domains. Throughout the table we use the symbols: \variable{a} is a robot arm, \variable{c} is a robot contact (fingertips or palm), \variable{o} is an object, \variable{p_{o}} is a pose of object \variable{o}, \variable{g_{o}} is a grasp on object \variable{o},  \variable{p_{ij}} is a relative pose between two objects \variable{i} and \variable{j}, \variable{q_{i}} is a configuration, \variable{r} is a region, \variable{t} is a trajectory, \variable{w} is a wrench, \variable{v} is a vise.} \label{table:fm}
\end{table*}


\begin{table*}[t!]
\begin{subtable}{\textwidth}
 \centering 
 \begin{tabular}{l c c}
   \hline
   Action & Preconditions & Effects \\ [0.5ex]
   \hline\hline
   grasp\_twist & 
   \makecell[cc]{(AtConf $a~q_{0}$) $\wedge$ (HandEmpty $a$) $\wedge$ (Cap $o_{c}$) $\wedge$ (Bottle $o_{b}$) $\wedge$ \\
                 (AtPose $o_{c}~p_{c}$) $\wedge$ (AtPose $o_{b}~p_{b}$)  $\wedge$ 
                 (Fixtured $o_{b}~w$) $\wedge$ \\
                 \textcolor{samplerGreen}{(StableGrasp $o_{c}~g_{c}~w$)} $\wedge$ 
                 \textcolor{samplerGreen}{(StableJoints $a~t~w$)} $\wedge$ \\
                 \textcolor{samplerGreen}{(GraspPushMotion $a~o_{c}~o_{b}~p_{c}~p_{b}~g_{c}~w~q_{0}~q_{1}~t$)} $\wedge$ \\
                  ($\neg$ (TrajUnsafe $a~t$))}  & 
   \makecell[cc]{(Twisted $o_{c}~w$) $\wedge$ \\ ($\neg$ (AtConf $a~q_{0}$)) $\wedge$ \\ (AtConf $a~q_{1}$)} \\
   \hline 
   contact\_twist & 
   \makecell[cc]{(AtConf $a~q_{0}$) $\wedge$ (HandEmpty $a$) $\wedge$ (Cap $o_{c}$) $\wedge$ (Bottle $o_{b}$) $\wedge$ 
                 (RContact $n$) $\wedge$ \\
                 (AtPose $o_{c}~p_{c}$) $\wedge$ (AtPose $o_{b}~p_{b}$) $\wedge$  
                 \textcolor{samplerGreen}{(ExtraWrench $w~w_{e}$)} $\wedge$  \\
                 (Fixtured $o_{b}~w_{e}$) $\wedge$ 
                 \textcolor{samplerGreen}{(StableJoints $a~t~w$)} $\wedge$ \\
                 \textcolor{samplerGreen}{(ContactPushMotion $a~n~o_{c}~o_{b}~p_{c}~p_{b}~w_{e}~q_{0}~q_{1}~t$)} $\wedge$ \\
                  ($\neg$ (TrajUnsafe $a~t$))} & 
   \makecell[cc]{(Twisted $o_{c}~w$) $\wedge$ \\ ($\neg$ (AtConf $a~q_{0}$)) $\wedge$ \\ (AtConf $a~q_{1}$)} \\
   \hline
   tool\_twist & 
   \makecell[cc]{(AtConf $a~q_{0}$) $\wedge$ (Cap $o_{c}$) $\wedge$ (Bottle $o_{b}$) $\wedge$ (PushTool $o_{t}$) $\wedge$ \\
                 (AtPose $o_{c}~p_{c}$) $\wedge$ (AtPose $o_{b}~p_{b}$) $\wedge$ (AtGrasp $a~o_{t}~g_{t}$) $\wedge$ \\ 
                 \textcolor{samplerGreen}{(ExtraWrench $w~w_{e}$)} $\wedge$ 
                 (Fixtured $o_{b}~w_{e}$) $\wedge$ \\
                 \textcolor{samplerGreen}{(StableGrasp $o_{t}~g_{t}~w_{e}$)} $\wedge$  
                 \textcolor{samplerGreen}{(StableJoints $a~t~w$)} $\wedge$ \\
                 \textcolor{samplerGreen}{(ToolPushMotion $a~o_{t}~o_{c}~o_{b}~p_{c}~p_{b}~g_{t}~w_{e}~q_{0}~q_{1}~t$)} $\wedge$ \\
                  ($\neg$ (TrajUnsafe $a~t$))} & 
   \makecell[cc]{(Twisted $o_{c}~w$) $\wedge$ \\ ($\neg$ (AtConf $a~q_{0}$)) $\wedge$ \\ (AtConf $a~q_{1}$)} \\
   \hline
   remove\_cap & 
   \makecell[cc]{(AtConf $a~q$) $\wedge$ (HandEmpty $a$) $\wedge$ (Cap $o_{c}$) $\wedge$ (Bottle $o_{b}$) $\wedge$ \\
                 (AtPose $o_{b}~p_{b}$) $\wedge$ (AtPose $o_{c}~p_{c}$) $\wedge$ (Twisted $o_{c}~w$) $\wedge$ \\
                 \textcolor{samplerGreen}{(UncapMotion $a~o_{c}~o_{b}~p_{b}~g~q~t$)} $\wedge$ \\
                  ($\neg$ (TrajUnsafe $a~t$))} & 
   \makecell[cc]{(AtGrasp $a~o_{c}~g$) $\wedge$ \\ ($\neg$ (AtPose $o_{c}~p_{c}$)) $\wedge$ \\ 
                 ($\neg$ (HandEmpty $a$)) $\wedge$ \\ (Uncapped $o_{c}~o_{b}$)} \\
 \end{tabular}
\caption{Actions} \label{table:bottleActions}
\end{subtable}
\vspace{0.5cm}
\begin{subtable}{\textwidth}
 \centering
 \begin{tabular}{l l}
   \hline
   Derived Fact & Definition \\ 
   \hline\hline
   (Fixtured $o~w$) & ((HoldingFixtured $o~w$) $\vee$ (SurfaceFixtured $o~w$) $\vee$ (ViseFixtured $o~w$)) \\
 \end{tabular}
\caption{Derived Facts} \label{table:bottleDerived}
\end{subtable}
\vspace{0.5cm}
\begin{subtable}{\textwidth}
 \centering
 \begin{tabular}{l c c l}
   \hline
   Sampler & Inputs & Outputs & Certified Facts \\
   \hline\hline
   sample-wrench & 
     $w_{1}$ & 
     $w_{2}$ & 
     \makecell[cc]{\textcolor{samplerGreen}{(ExtraWrench $w_{1}~w_{2}$) $\wedge$ (Wrench $w_{2}$)}}\\
   \hline
   plan-uncap-motion & 
     $a~o_{c}~o_{b}~p_{b}~g_{c}$ & 
     $q~t$ & 
     \makecell[cc]{\textcolor{samplerGreen}{(UncapMotion $a~o_{c}~o_{b}~p_{b}~g_{c}~q~t$) $\wedge$ (Conf $q$) $\wedge$ (Traj $t$)}} \\
   \hline
   plan-grasp-twist & 
     $a~o_{c}~o_{b}~p_{c}~p_{b}~g_{c}~w$ & 
     $q_{0}~q_{1}~t$ & 
     \makecell[cc]{\textcolor{samplerGreen}{(GraspPushMotion $a~o_{c}~o_{b}~p_{c}~p_{b}~g_{c}~w~q_{0}~q_{1}~t$) $\wedge$} \\ 
                   \textcolor{samplerGreen}{(Conf $q_{0}$) $\wedge$ (Conf $q_{1}$) $\wedge$ (Traj $t$)}} \\
   \hline
   plan-contact-twist & 
   $a~n~o_{c}~o_{b}~p_{c}~p_{b}~w$ & 
   $q_{0}~q_{1}~t$ & 
   \makecell[cc]{\textcolor{samplerGreen}{(ContactPushMotion $a~n~o_{c}~o_{b}~p_{c}~p_{b}~w~q_{0}~q_{1}~t$) $\wedge$} \\ 
                 \textcolor{samplerGreen}{(Conf $q_{0}$) $\wedge$ (Conf $q_{1}$) $\wedge$ (Traj $t$)}} \\
   \hline
   plan-tool-twist & 
   $a~o_{t}~o_{c}~o_{b}~g_{t}~p_{c}~p_{b}~w$ & 
   $q_{0}~q_{1}~t$ & 
   \makecell[cc]{\textcolor{samplerGreen}{(ToolPushMotion $a~o_{t}~o_{c}~o_{b}~p_{c}~p_{b}~g_{t}~w~q_{0}~q_{1}~t$)} $\wedge$ \\ 
                 \textcolor{samplerGreen}{(Conf $q_{0}$) $\wedge$ (Conf $q_{1}$) $\wedge$ (Traj $t$)}} \\
 \end{tabular}
\caption{Samplers} \label{table:bottleSamplers}
\end{subtable}
\caption{The domain-specific lifted operators, derived facts and samplers for the childproof bottle domain. Throughout the table we use the symbols: \variable{a} is a robot arm, \variable{o} is an object, \variable{p_{o}} is a pose of object \variable{o}, \variable{g_{o}} is a grasp on object \variable{o}, \variable{q_{i}} is a configuration, \variable{r} is a region, \variable{t} is a trajectory, \variable{w} is a wrench. Specific to this domain: \variable{n}, \variable{o_{b}}, \variable{o_{c}} and \variable{o_{t}} refer to the robot contact, bottle, cap and pusher tool, respectively. } \label{table:bottle}
\end{table*}


\begin{table*}[t!]
\begin{subtable}{\textwidth}
 \centering
 \begin{tabular}{l c c}
   \hline
   Action & Preconditions & Effects \\ [0.5ex]
   \hline\hline 
   slice\_cut & \makecell[cc]{(AtConf $a~q_{0}$) $\wedge$ (Cuttable $o_{v}$) 
                        $\wedge$ (AtPose $o_{v}~p_{v}$) $\wedge$ (Knife $o_{k}$) $\wedge$ \\
                        (AtGrasp $a~o_{k}~g_{k}$) $\wedge$ 
                        (Fixtured $o_{v}~w_{0}$) $\wedge$ (Fixtured $o_{v}~w_{1}$) $\wedge$ \\
                        \textcolor{samplerGreen}{(StableGrasp $o_{k}~g_{k}~w_{0}$)} $\wedge$ 
                        \textcolor{samplerGreen}{(StableGrasp $o_{k}~g_{k}~w_{1}$)} $\wedge$ \\
                        \textcolor{samplerGreen}{(StableJoints $a~t~w_{0}$)} $\wedge$ 
                        \textcolor{samplerGreen}{(StableJoints $a~t~w_{1}$)} $\wedge$ \\
                        \textcolor{samplerGreen}{(SliceCutMotion $a~o_{k}~o_{v}~g_{k}~p_{c}~w_{1}~w_{2}~q_{0}~q_{1}~t$)}$\wedge$ \\
                        ($\neg$ (TrajUnsafe $a~t$))} & 
   \makecell[cc]{(Sliced $o_{v}$) $\wedge$ \\ ($\neg$ (AtConf $a~q_{0}$)) $\wedge$ \\ (AtConf $a~q_{1}$)} \\
 \end{tabular}
\caption{Actions} \label{table:cutActions}
\end{subtable}
\vspace{0.5cm}
\begin{subtable}{\textwidth}
 \centering
 \begin{tabular}{l l}
   \hline
   Derived Fact & Definition \\ 
   \hline\hline
   (Fixtured $o~w$) & ((HoldingFixtured $o~w$) $\vee$ (SurfaceFixtured $o~w$) $\vee$ (ViseFixtured $o~w$)) \\
 \end{tabular}
\caption{Derived Facts} \label{table:cutDerived}
\end{subtable}
\vspace{0.5cm}
\begin{subtable}{\textwidth}
 \centering
 \begin{tabular}{l c c l}
   \hline
   Sampler & Inputs & Outputs & Certified Facts \\
   \hline\hline
   plan-slice-motion & 
      $a~o_{k}~o_{v}~g_{k}~p_{v}~w_{1}~w_{2}$ & 
      $q_{0}~q_{1}~t$ & 
      \makecell[cc]{\textcolor{samplerGreen}{(SliceCutMotion $a~o_{k}~o_{v}~g_{k}~p_{v}~w_{1}~w_{2}~q_{0}~q_{1}~t$) $\wedge$} \\ 
                    \textcolor{samplerGreen}{(Conf $q_{0}$) $\wedge$ (Conf $q_{1}$) $\wedge$ (Traj $t$)}} \\ 
 \end{tabular}
\caption{Samplers} \label{table:cutSamplers}
\end{subtable}
\caption{The domain-specific lifted operators, derived facts and samplers for the vegetable cutting domain. Throughout the table we use the symbols: \variable{a} is a robot arm, \variable{o} is an object, \variable{p_{o}} is a pose of object \variable{o}, \variable{g_{o}} is a grasp on object \variable{o}, \variable{q_{i}} is a configuration, \variable{r} is a region, \variable{t} is a trajectory, \variable{w} is a wrench. Specific to this domain: \variable{o_{k}} and \variable{o_{v}} refer to the knife and the vegetable, respectively.} \label{table:cut}
\end{table*}

In \sref{sec:fkcPDDL} and \sref{sec:fixturePDDL} we detailed the augmentations needed to enable forceful manipulation in the PDDLStream framework.
In that example we discussed how the forceful kinematic chain constraint was integrated into the domain and discussed two possible fixturing methods: \texttt{(HoldingFixtured \variable{o}~\variable{w})}, which uses a second robot to grasp the object to be fixtured, and \texttt{(WeightFixtured \variable{o}~\variable{w})}, which fixtures by weighing down the object with a heavy mass. 

\tref{table:fm} lists the lifted operators, derived facts and samplers that are used to enable the other fixturing strategies that can be used across forceful manipulation domains: \texttt{(ViseFixtured \variable{o}~\variable{w})} and \texttt{(SurfaceFixtured \variable{o}~\variable{w})}

One way to fixture an object is to place it in a vise: \texttt{(ViseFixtured \variable{o}~\variable{w})}. 
This fact is verified by the test sampler \texttt{test-vise-grasp-stable}, which uses the limit surface models to evaluate the stability of the grasp. 
To use a vise we add the lifted operators \texttt{open\_vise} and \texttt{close\_vise} along with their sampler \texttt{plan-vise-motion}. 
These operators actuate the vise (implemented here as a table-mounted robotic hand) and have as a precondition that the object to be fixtured is already in the vise, which can be achieved through a pick-and-place motion. 

Another way to fixture is to create a frictional grasp by exerting a downward force on the object through a contact, thus sandwiching the object between the contact and the surface \texttt{(SurfaceFixtured \variable{o}~\variable{w})}.
The exerted downward force can be thought of as the grasping force of the frictional grasp. 
This creates two joints: the joint between the force-exerting contact and the object (evaluated by \texttt{test-contact-stable}) and the joint between the object and the surface (evaluated by \texttt{test-surface-stable}). 
For both joints we use the limit surface models to evaluate stability.

We also introduce operators to make and break contact with either a grasped tool or an end effector contact of the robot (fingers or palm): \texttt{pushin\_tool}, \texttt{pushout\_tool}, \texttt{pushin\_rcontact}, \texttt{pushout\_rcontact} (with samplers \texttt{plan-tool-contact} and \texttt{plan-rcontact-contact}).
The operators that make contact (\texttt{pushin\_tool} and \texttt{pushin\_rcontact}) make contact by exert a downward force on the object, leveraging Cartesian Impedance control, described in \sref{appendix:controller}.

In each domain we define a derived fact \texttt{(Fixtured \variable{o}~\variable{w})} to define which fixturing methods are valid to use. 


\subsection{Childproof Bottle Domain}

\tref{table:bottle} lists the lifted operators, derived facts and samplers that are specific to the childproof bottle domain.
The robot can impart the forceful operation to push-twist the cap through a variety of possible contacts: a grasp, fingertips, a palm or a grasped pusher tool.
Correspondingly, we have a lifted operator for each of these contact types: \texttt{grasp\_twist}, \texttt{contact\_twist}, \texttt{tool\_twist}. 
The operator \texttt{contact\_twist} is parameterized by what end effector contact the robot is using, which we define as being either the fingertips or a palm. 

The controller for the \texttt{grasp\_twist} operator grasps the cap with the hand, forcefully pushes down while twisting the cap with the hand and then releases the cap. 
The controller for the \texttt{contact\_twist} operator performs the same sequence only instead of grasping the cap, frictional planar contact is made. 
Likewise the controller for the \texttt{tool\_twist} operator is the same sequence, but where a grasped tool is making and breaking contact with the cap. 
The trajectories are generated by the samplers \texttt{plan-grasp-twist}, \texttt{plan-contact-twist} and \texttt{plan-tool-twist} respectively. 
Unsurprisingly, the implementation of the three samplers has a tremendous amount of overlap with small, parameterized differences. 
We use a Cartesian Impedance controller, detailed in \sref{appendix:controller} to exert the forceful push and twist wrench for all three operators.

While the task requires that the robot exert a wrench $w=(0, 0, -f_{z}, 0, 0, t_{z})$, we allow \texttt{contact\_twist} and \texttt{tool\_twist} to sample a wrench $w_{e}$ that exerts additional downward force, i.e. $w_{e}=(0, 0, -f_{z++}, 0, 0, t_{z})$ where $f_{z++}$ is a sampled value such that $f_{z++} > f_{z}$.
If the bottle is being fixtured via \texttt{(SurfaceFixtured \variable{o}~\variable{w})} this additional force increase the stability of the frictional contacts. 


Like in the nut twisting domain, any operator that applies a forceful operation has preconditions that check the forceful kinematic chain constraint. 
Note that for the operators where the robot is exerting \variable{w_{e}} the stability is evaluated with respect to this wrench. 

For each of these operators, the bottle must be fixtured, which can either be achieved by using a second arm to grasp the bottle (\texttt{(HoldingFixtured \variable{o}~\variable{w})}), using the frictional contact with the surface (\texttt{(SurfaceFixtured \variable{o}~\variable{w})}) or using a vise (\texttt{(ViseFixtured \variable{o}~\variable{w})}). 
Likewise if the operator is exerting \variable{w_{e}} the fixturing is evaluated with respect to this. 
In this domain we slightly modify the formulation of \texttt{(SurfaceFixtured \variable{o}~\variable{w})} to account for the fact that the additional force is made in conjuction with the push-twist operations. 

We additionally add a lifted operator that removes the cap from the bottle, \texttt{remove\_cap}, which is only feasible after we have used one of the push-twist operators.
The controller for this operator grasps the cap with the hand and lifts the cap up. 


\subsection{Vegetable Cutting Domain}

\tref{table:cut} lists the lifted operator, derived fact and sampler that are specific to the vegetable cutting domain.
The robot imparts the forceful operation to cut the vegetable by using a grasped knife.

We define one new operator: \texttt{slice\_cut}. 
The controller for this operator makes contact with the vegetable via the grasped knife, exerts a downward force with the knife and then exerts a translational slice with the knife. 
These trajectories are generated by sampler \texttt{plan-slice-motion} and, like in the other domains, we use Cartesian Impedance control to exert the wrenches.

Since this operator involves two wrenches, the wrench of the downward force $w_{0}$ and the wrench of the translation slice $w_{1}$, the forceful kinematic chains must be stable with respect to both wrenches. 
Hence the preconditions evaluate the stability of the vegetable's fixturing and the stability of the grasp on the knife for $w_{0}$ and $w_{1}$.  

In this domain the vegetable must be fixtured, which can be achieved either by using a second arm to grasp the vegetable (\texttt{(HoldingFixtured \variable{o}~\variable{w}}), using the frictional contact with the surface (\texttt{(SurfaceFixtured \variable{o}~\variable{w})}) or using a vise (\texttt{(ViseFixtured \variable{o}~\variable{w})}.

\end{document}